\documentclass[acmsmall,screen]{acmart}

\setcopyright{rightsretained}
\acmJournal{PACMHCI}
\acmYear{2024} \acmVolume{8} \acmNumber{CSCW1} \acmArticle{85} \acmMonth{4} \acmPrice{}\acmDOI{10.1145/3637362}

\usepackage{graphicx} 
\usepackage{natbib}  
\usepackage{caption} 
\usepackage[most]{tcolorbox}
\usepackage{colortbl}
\usepackage{multirow}
\usepackage{adjustbox}
\usepackage{amsmath,latexsym}
\usepackage{amsfonts}
\usepackage{amssymb}
\usepackage{graphicx}
\usepackage{amsmath,latexsym}

\usepackage{amsfonts}
\usepackage{amssymb}
\usepackage{paralist}
\usepackage{xspace}
\usepackage{hyperref}
\usepackage{booktabs} 
\usepackage{wasysym} 
\usepackage{array}
\usepackage{multirow}
\usepackage{amssymb}
\usepackage{breqn}
\usepackage{amsfonts}
\usepackage{amssymb}
\usepackage{paralist}
\usepackage{xspace}
\usepackage{hyperref}
\usepackage{booktabs} 
\usepackage{graphicx}
\usepackage{subfig}

\usepackage{wasysym} 
\usepackage{array}
\usepackage{multirow}
\usepackage{amssymb}
\usepackage{breqn}

\usepackage{tikz}
\usepackage{pgfplots}
\usepackage{float}
\usepackage{arydshln}
\usepackage{centernot}
\usepackage{bbm}
\usepackage{enumitem}
\usepackage{algorithm}
\usepackage{algorithmic}
\usepackage{enumitem}
\usepackage[normalem]{ulem}
\usepackage{tabulary}
\usepackage{subfig}
\usepackage[mathscr]{euscript}
\usepackage{multirow}

\usepackage{amsmath,latexsym}
\usepackage{amsfonts}
\usepackage{amssymb}

\newcommand{\revision}[1]{\textcolor{black}{#1}}

\hypersetup{
	colorlinks=true,       
	linkcolor=black,        
	citecolor=black,        
	filecolor=black,        
	urlcolor=black         
}
\usepackage{array}
\usepackage{makecell}
\usepackage{multirow}
\usepackage{newfloat}
\usepackage{listings}

\usepackage{booktabs}
\usepackage{longtable}
\pgfplotsset{compat=1.18}

\begin{document}

\title{Improving User Behavior Prediction: Leveraging Annotator Metadata in Supervised Machine Learning Models}


\author{Lynnette Hui Xian Ng}
\affiliation{%
 \institution{Carnegie Mellon University}
 \country{USA}}

\author{Kokil Jaidka}
\affiliation{%
  \institution{NUS Centre for Trusted Internet \& Community, National University of Singapore}
  \streetaddress{11 Computing Drive}
  \city{Singapore}
  \country{Singapore}}
\email{jaidka@nus.edu.sg}

\author{Kaiyuan Tay}
\affiliation{%
 \institution{National University of Singapore}
 \country{Singapore}}

\author{Hansin Ahuja}
\affiliation{%
  \institution{Indian Institute of Technology Ropar}
  \city{Ropar}
  \country{India}
}

\author{Niyati Chhaya}
\affiliation{%
 \institution{Adobe Research}
 \country{India}}

\renewcommand{\shortauthors}{Ng et. al.}

\begin{abstract}
    Supervised machine-learning models often underperform in predicting user behaviors from conversational text, hindered by poor crowdsourced label quality and low NLP task accuracy. We introduce the Metadata-Sensitive Weighted-Encoding Ensemble Model (MSWEEM), which integrates annotator meta-features like fatigue and speeding. First, our results show MSWEEM outperforms standard ensembles by 14\% on held-out data and 12\% on an alternative dataset. Second, we find that incorporating signals of annotator behavior, such as speed and fatigue, significantly boosts model performance. Third, we find that annotators with higher qualifications, such as Master's, deliver more consistent and faster annotations. Given the increasing uncertainty over annotation quality, our experiments show that understanding annotator patterns is crucial for enhancing model accuracy in user behavior prediction.

\end{abstract}

\maketitle
\section{Introduction}
Natural language processing (NLP) tasks aimed at predicting user behavior typically involve analyzing text snippets authored by individuals. These tasks require annotations that reflect various behavioral outcomes, such as betrayal~\cite{niculae2015linguistic,peskov-etal-2020-takes}, persuasion~\cite{yang2019let}, or editing behavior~\cite{jaidka2021wikitalkedit}. Consequently, these tasks are formulated as text classification problems where supervised machine learning models analyze the text and predict behavioral labels. These texts, often sourced from social media interactions, indicate user behavior in social contexts. However, behavioral-based NLP tasks tend to have lower prediction accuracy compared to other text classification tasks due to two significant challenges: the need to incorporate semantic knowledge into model training and ensuring high-quality label annotation~\cite{hovy2021importance,yang2019let}.

Our approach is motivated by Kahneman's theory of attention~\cite{kahneman1973attention}, which posits that attention is a limited cognitive resource allocated dynamically based on task demands and individual effort. Tasks requiring significant mental effort or prolonged engagement deplete attention, potentially leading to reduced output quality. Conversely, tasks that are overly automated or repetitive may fail to engage sufficient attention, compromising task performance. Drawing on this framework, we consider the effect of annotation quality on machine learning by modeling annotator behavior related to attention and effort. Specifically, we have derived \textit{worktime} to measure speeding and \textit{throughput} to assess fatigue. These features provide critical insights into annotator performance, enabling strategies to improve label quality by addressing cognitive constraints during annotation.

Incorporating semantic knowledge into model training is also crucial for enhancing prediction accuracy. {Ensemble learning approaches, which combine predictions from multiple models, address performance variance and improve prediction reliability by leveraging diverse insights into the domain~\cite{brownlee2018better,xu2015argumentation,PERIKOS2016191,zimmerman2018improving,chen2021weakly}. Ensemble learning frameworks typically re-imagine a multilabel classification problem by combining predictions about facets of the final outcome~\cite{xu2015argumentation,PERIKOS2016191,zimmerman2018improving,ng2023botbuster}. By assembling multiple base learners specific to various facets of a task, ensemble architectures achieve better performance and generalization~\cite{yang2023survey}. Frameworks like these contribute to richer insights into linguistic strategies predictive of behavioral outcomes. Studies have validated ensemble learning frameworks for text classification problems, including disclosure~\cite{xincl}, argumentation~\cite{xu2015argumentation}, hate speech~\cite{zimmerman2018improving,parikh2019multi}, spam detection~\cite{saeed2022ensemble}, and detection of automated texts in social media~\cite{ng2023botbuster}. However, there remains uncertainty in linking textual properties to user behavior, particularly in behavioral-based tasks.

The quality of crowdsourced labels poses another critical challenge. Advances in large language models (LLMs) have raised concerns about the integrity of annotation tasks. Recent research indicates that LLM-supported annotations can improve but may not always outperform crowdsourced annotations in identifying ground truth~\cite{he2024if}. Persistent issues in crowdsourced tasks, such as inattentiveness, multitasking~\cite{saravanos2021hidden,litman2015relationship,chandler2014nonnaivete}, and reliance on bots or scripts~\cite{douglas2023data,veselovsky2023artificial}, exacerbate these challenges. Annotator bias, lack of expertise~\cite{hsueh2009data,bayerl2011determines,fornaciari2021beyond}, and the variability of annotator behaviors further highlight the need for reliable strategies to enhance data quality.

Our work addresses these challenges by proposing a weighted ensemble framework for machine learning. We enrich this framework with annotator-specific meta-features—\textit{worktime}, \textit{throughput}, \textit{percentage agreement}, and \textit{text length}—that capture behavioral and task-specific factors during annotation. These features, detailed in~\autoref{sec:metafeatures}, are collected during annotation and calibrated to the unique needs of various cohorts. While our experiments primarily utilize Amazon Mechanical Turk, the principles generalize to other crowdsourcing platforms focused on human-labeled text classification. Even platforms lacking direct annotator performance signals can estimate these features (e.g., worktime, throughput), ensuring practical applicability.

Our contributions are as follows: \begin{itemize}[noitemsep] \item We propose a novel set of annotator meta-features, including \textit{worktime} and \textit{throughput}, to extend prior work that primarily relied on \textit{percentage agreement} and \textit{text length} for addressing label quality. \item We validate these features through extensive experiments, demonstrating their efficacy in training ensemble models for diverse datasets and tasks. \item We offer insights into annotator qualifications and behaviors, providing actionable guidance for designing future crowdsourcing tasks. \end{itemize}

Our experiments leverage the CLAff-Diplomacy, CLAff-OffMyChest, and Counterfire datasets, which provide annotation metadata to model annotator behavior as a function of the text prompts provided. The CLAff-Diplomacy dataset~\cite{jaidka2020report} models the anticipation of deception during conversations, with auxiliary annotations related to message semantics. The CLAff-OffMyChest dataset~\cite{DBLP:conf/aaai/JaidkaSLCU20} focuses on the popularity of posts, offering annotations for emotional and informational disclosure. The Counterfire dataset~\cite{verma2024auditing} includes metadata from annotators assessing the linguistic quality of social media posts.

\section{Background}
\label{sec:background}
Annotations for most tasks are obtained through platforms like Prolific, Appen, CrowdSource, and Amazon Mechanical Turk (AMT). On AMT, annotators known as MTurkers assign a label to a text presented. These crowdsourced workers are paid according to the number of text annotation assignments they complete. A past study suggests that employing expert annotators with theoretical and applied knowledge of the domain leads to better annotations~\cite {waseem-2016-racist}, and by extension, better models than crowdsourced amateur annotators. However, in many cases, expert annotators are unavailable or are expensive to engage, or the annotations have already been collected. Yet, the suboptimal conditions involved in crowdsourcing annotations from non-experts is often ignored in research analyses. 

Crowdsourcing allows for monitoring annotator speed but requires additional measures beyond completion time to ensure quality. Annotators might produce low-quality work due to overwork, multitasking, or fatigue despite meeting time thresholds. Our approach is inspired by Kahneman’s theory of attention, as discussed in \textit{Attention and Effort}~\cite{kahneman1973attention}. Kahneman posits that attention is a limited cognitive resource allocated dynamically based on task demands and individual motivation. Tasks requiring significant mental effort, such as complex problem-solving, deplete attention more quickly, while routine or automated tasks demand less cognitive effort. This duality explains why individuals can switch between effortful and effortless processing, balancing cognitive load and performance. Kahneman also highlights how factors like task complexity, fatigue, and environmental cues influence attention distribution and the quality of task execution.
Drawing upon this theory, our work leverages meta-features such as \textit{Worktime}, \textit{Throughput}, and \textit{Text Length} to model annotators’ attention and effort during task performance. These meta-features are motivated by prior work in crowdsourced annotation, which has examined behavioral and linguistic indicators of annotation quality~\cite{hovy2010towards, chen2022construction, hsueh-etal-2009-data, saeed2018impact}. For example, longer worktimes may signal increased attention allocation, while extremely short durations may indicate insufficient cognitive engagement or speeding~\cite{eickhoff2013increasing}. Similarly, throughput provides insights into task repetition and efficiency, with higher throughput potentially signaling habituation or declining attention over time~\cite{difallah2018demographics}. Text length, a measure of task complexity, can further influence annotator fatigue, impacting annotation quality~\cite{kazai2011worker}. These meta-features align with prior research in crowdsourcing and human performance evaluation and complement alternative approaches such as weighting schemes based on label similarity~\cite{tao2020label}, annotator expertise inference~\cite{NIPS2009_f899139d}, and noise detection functions~\cite{hsueh-etal-2009-data}.

Our work complements prior research that explores additional meta-features beyond inter-annotator agreement. For example, probabilistic models have incorporated annotator-specific error functions into expectation-maximization~\citep{ipeirotis2010quality}, while domain-specific weighting has been explored for label aggregation~\cite{tao2018domain}. More recently, data-driven approaches have evaluated annotation quality by benchmarking model performance on individual annotations~\citep{sedoc2020item}. Furthermore, efforts to account for annotator behavior in machine learning models have been explored in work by Geva et al.~\cite{geva2019we}, who highlight how certain annotation behaviors can bias model training and lead to over-estimation. We extend their idea by incorporating behavioral meta-features and traditional quality markers into our modeling framework.

Additionally, we benchmark our work against established approaches that integrate annotator metadata. Our first baseline, MACE (Multi Annotator Competence Estimate), applies a generative Bayesian probability model to estimate annotator credibility~\citep{hovy-etal-2013-learning}. Other approaches have leveraged clustering techniques to refine label aggregation~\cite{CHATTERJEE2017138}. By incorporating diverse meta-features informed by prior research, we aim to provide a systematic approach to analyzing annotator performance and improving annotation quality.

\subsection{Integrating annotator signals of quality for crowdsourced labels}
Most researchers use pairwise or average agreement to threshold data quality to an acceptable minimum when collecting text annotations on crowdsourced platforms like AMT. Strategies to maintain high-quality annotations include employing multiple independent annotators per data point~\cite{hovy2010towards,chen2022construction}, selecting and training annotators effectively~\cite{carlson2003building}, sampling data representatively~\cite{chen2022construction}, and optimizing annotation sequence~\cite{lourentzou2019data}. Prior work has mainly focused on annotation inconsistency as the \textit{only} facet of annotation quality, which we think is an erroneous assumption. Rodrigues and Pereira~\cite{rodrigues2018deep} weighed observations based on inter-annotator agreement (which is only one part of our approach). Other approaches that involve majority voting have also proposed weighting schemes focused on domain~\cite{tao2018domain}, label similarity~\cite{tao2020label}, probabilistic inference of the annotator expertise~\cite{NIPS2009_f899139d}, or clustering individual annotations~\cite{CHATTERJEE2017138} to derive the target variable. More advanced techniques have used quality-sensitive errors to represent annotator reliability. For example, previous work~\citep{ipeirotis2010quality} built in a probabilistic error into the expectation-maximization step, while another study~\citep{rodrigues2018deep} incorporated annotation-specific backpropagation errors in their training framework. Techniques proposed to eliminate data produced by non-expert annotators include identifying low-quality data through an accumulated noise function~\cite{hsueh-etal-2009-data} or via linguistic properties of written annotations~\cite{saeed2018impact}. Both methods led to an increase in prediction accuracy for the downstream task. Finally, prior work has also evaluated data quality based on model performance on individual observations~\citep{sedoc2020item}. These approaches use a benchmark for reference and involve the contentious decision of dropping or revising observations based on model performance. 

Our efforts complement a prior study that has reported how other annotator behaviors can affect label quality and lead to model over-estimation issues~\cite{geva2019we}. We extend their idea by adding meta-features to the machine learning model with noise labels. We also benchmark our work against the two recent studies that factor annotator metadata into their approaches. Our first baseline is MACE (Multi Annotator Competence Estimate)~\citep{hovy-etal-2013-learning}, which uses a generative Bayesian probability method to model annotator credibility. 
Our second baseline replicates combinatorial approaches that involve inter-annotator agreement and error patterns in its model training~\cite{geva2019we}. However, both our baselines ignore the impact of annotator fatigue or any other signals of macro-level annotator behavior, such as their speed or throughput, on the downstream text classifier performance. The baselines are also only suitable for generating better quality labels for one set of annotations at a time but are not applicable for inferring higher quality labels for an ensemble setup.
\section{Method}


We propose to improve the performance of models used to predict user behavior with ensemble learning frameworks that incorporate semantic knowledge and address label quality issues. Our approach involves integrating annotator meta-features in a machine-learning ensemble to improve text classification performance. We pose the following questions:
\begin{itemize}[noitemsep]
    \item What is the effect of enriching ensemble-based text classification approaches with metadata about the annotations?
    \item How well does this approach generalize to new data with different auxiliary and primary tasks?
\end{itemize}

Specifically, we define the following problem statement:
Given a series of input data, behavioral meta-data features, and ground-truth outputs, train a Metadata-Sensitive Weighted-Encoding Ensemble Model (MSWEEM) that:
\begin{itemize}[noitemsep]
    \item Leverages inherent textual properties in an ensemble design to predict a \textit{target variable}. The \textit{target variable} is the final variable that we want to predict. The inherent textual properties are signals of the target variable, based on the domain expertise of the researchers. These properties are represented as \textit{auxiliary variables} in the model.
    \item Enriches an ensemble machine-learning model with behavioral metadata information as a probabilistic encoding of latent annotation variables. The latent annotation variables are coded as \textit{meta-features} to provide additional information for prediction.
    \item Generalizes to other datasets with different primary and auxiliary prediction tasks but similar meta-features.
\end{itemize}

Our ensemble uses behavioral data from crowdsourced annotations for each sub-label as a weighting scheme before aggregating the labels to get a final label. Specifically, the ensemble models are enriched with four facets of annotator behavior: speeding (worktime), fatigue (throughput), percentage agreement, and text length, collected at the time of annotation. These facets are detailed in~\autoref{sec:metafeatures}. We validate our model's generalizability across two datasets: CLAff-OffMyChest and CLAff-Diplomacy\footnote{Our code and data are available at \url{https://anonymous.4open.science/r/diplomacy-betrayal-E454/README.md}}.
\footnote{Our code and data are available at https://github.com/quarbby/diplomacy-betrayal}. 
We show that our framework generalizes to different text classification tasks within the same domain by retraining the method for deriving predictions from the penultimate neural layer, using techniques such as majority voting~\cite{xu2015argumentation,PERIKOS2016191} or max pooling~\cite{zimmerman2018improving}.

Our experiments leverage the openly available CLAff-Diplomacy and CLAff-OffMyChest datasets. These datasets model user behavior based on text prompts provided to users. The CLAff-Diplomacy dataset~\cite{jaidka2020report} targets the receiver's anticipation of deception in conversations, offering auxiliary annotations on the semantic qualities of messages exchanged during a game. Conversely, the CLAff-OffMyChest dataset targets post popularity~\cite{DBLP:conf/aaai/JaidkaSLCU20}, with auxiliary annotations on emotional and informational disclosure and support.

\subsection{Meta-features from Amazon Mechanical Turk}
\label{sec:metafeatures}
To examine the effects of annotator quality on text classification performance, we devised meta-features that would capture behavioral and task-specific factors during annotation and address key concerns such as speeding, fatigue, and task complexity. Specifically, we include the following meta-features:

Behavioral signals such as \textbf{Throughput (TP)} and \textbf{Worktime (WT)} that have been examined in previous studies: \begin{itemize} \item \textbf{Throughput (TP)}: Prior work identifies throughput as a critical metric for understanding annotator productivity~\cite{difallah2018demographics}. While it reflects worker engagement and efficiency, excessively high throughput may indicate potential quality issues, aligning with findings on annotation inconsistency and noise filtering~\cite{hsueh-etal-2009-data}. \item \textbf{Worktime (WT)}: Task duration has been used to detect speeding behavior, which can compromise annotation quality. For example, Eickhoff and de Vries (2013) demonstrated that task duration could help identify fraudulent or inattentive behaviors in crowdsourced tasks~\cite{eickhoff2013increasing}. Worktime has also been considered in models estimating annotator expertise~\cite{NIPS2009_f899139d}. \end{itemize} 

Traditional quality measures, such as \textbf{Percentage Agreement (PC)} and \textbf{Text Length (TL)}, further complement the analysis: \begin{itemize} \item \textbf{Percentage Agreement (PC)}: Agreement between annotators is a widely used measure of inter-rater reliability~\cite{artstein2008inter}. Artstein and Poesio (2008) discuss evaluation metrics such as Cohen’s kappa and Krippendorff’s alpha, which assess the consistency of annotations. While most approaches emphasize inter-annotator agreement as a primary quality measure~\cite{rodrigues2018deep}, we argue that it is only one facet of annotation quality. \item \textbf{Text Length (TL)}: Text length features, such as word and character counts, are indicative of task complexity and can signal task engagement or annotator fatigue~\cite{kazai2011worker}. Previous studies have also found that linguistic properties of text can serve as indicators of annotation reliability~\cite{saeed2018impact}. \end{itemize} 

These indicators, detailed in~\autoref{tab:metadatafeatures}, form the foundation of our analysis and serve as inputs for our ensemble learning framework. By leveraging these meta-features, our approach aims to enhance label quality and ensure robust model performance.

\begin{table}[!ht]
\caption{Metadata features $M$ for each observation. They are scaled using min-max normalization to a [0,1] distribution.}
\label{tab:metadatafeatures}
\begin{center}
\small
\begin{tabular}{p{3cm}p{5cm}p{5cm}}
\hline
\textbf{Feature} & \textbf{Description } & \textbf{Distribution} \\ 
&\footnotesize{\textbf{\textit{(No. of features)}}}&\\
\hline
\multicolumn{3}{c}{\textbf{Behavioral signals evaluated in this paper}}\\
\hline
Throughput (TP) &  \footnotesize{Total tasks completed \textit{(5)}} & M$_{TP}$ = 1271.9\newline M$_{TP}\in[$1,3747$]$ \\ 
\hline
Worktime (WT) & \footnotesize{Time taken to complete one task in seconds \textit{(5)}} & M$_{WT}$ = 297.4; \newline M$_{WT}\in[$5,10477$]$\\
\hline
\multicolumn{3}{c}{\textbf{Traditional markers of data quality}}\\
\hline
Percentage \newline Agreement (PC) & \footnotesize{\% Agreements for assigned labels \textit{(4)}} &  M$_{PC}$ = 80.2; \newline M$_{PC}\in[$60,100$]$ \\
\hline
Text Length (TL) &  \footnotesize{Number of words (w) and characters (c) \textit{(2)}} & M$_{w}$ = 16.5; \newline M$_{w}\in[$7,96$]$ \newline M$_{c}$ = 84.4; \newline M$_{c}\in[$24,505$]$ \\


\hline
\end{tabular}
\end{center}
\end{table}

\subsection{Metadata-Sensitive Weighted-Encoding Ensemble Model (MSWEEM)}
\label{sec:model}
We propose an ensemble approach from annotated label modeling to target user behavior prediction. Behavioral labels, which comprise our target variables, are derived from the actions or reactions of individuals. They depend on users' interpretation and interaction. The overall architecture of our \textbf{M}etadata-\textbf{S}ensitive \textbf{W}eighted-\textbf{E}ncoding \textbf{E}nsemble \textbf{M}odel \textbf{(MSWEEM)} is described in Figure \ref{fig:pipeline}. Given a corpus of $N$ sentences $S = \{s_i\}^N_{i=1}$, each sentence contains $P$ annotations, $a = \{a_j\}_{j=1}^P$ and $Q$ meta-features $M = \{M_l\}_{l=1}^Q$. We want to learn the target variable with the aid of semantic labels that are derived directly from the content's inherent attributes. The semantic attributes are defined as the `auxiliary variables,' and they aid in understanding the text to classify the target variable more accurately. 

Figure~\ref{fig:pipeline} demonstrates how conversational text is annotated for one target variable and four auxiliary variables. Each data point in our study is labeled for four auxiliary variables. Each auxiliary variable offers a binary classification problem. The classifier $\{f_{aj}\}$ of each label thus outputs a posterior probability distribution of each label $\{f_{aj}(x)\}$. 

We weigh each prediction proportionately to its prior distribution to account for dataset imbalance among these labels. This prior distribution is derived from the proportion of each label among the sub-dataset used for training. The weights of each label thus represents the prevalance of each label within the training dataset.

The four probability encodings produced by the individual models are then constructed into an encoding vector as an input to the ensemble layer $H$.\footnote{The details of the individual model construction are described in Section 5.2.} 
We reflect the correctness of the annotation of the auxiliary variables using the behavioral meta-features of the annotators. We do so by multiplying the encodings with the annotation metadata value $M_l$ to obtain $\{M_lp_{aj}\}$. Finally, these probabilities encoded with annotation metadata are passed through an activation layer to predict the target variable $y$. 

Throughout this paper, we use several notations to represent variables in our models.~\autoref{tab:notations} summarizes the notations we use in this paper.
\begin{table}[!ht]
\caption{\revision{Explainations of Variables and Noations}}
\label{tab:notations}
\begin{center}
\small
\begin{tabular}{lp{7cm}p{4cm}}
\hline
\textbf{Variable} & \textbf{Explanation} & \textbf{\revision{Shape}} \\ \hline
$M_l$ & Annotation metadata & \revision{float} \\
$s_i$ & Text sentence & \revision{string} \\
$a_j$ & \revision{Auxiliary variable, derived from} annotation labels \revision{that reflect} semantic properties \revision{of the text} & \revision{integer representing the class} \\
$y$ & Target variable denoting a behavioral outcome & \revision{integer representing the class} \\
$f_{aj}$ & Classifier of annotation label $a_j$ & \revision{function} \\
$p_{aj}$ & Probabilistic encodings of classifier $aj$ & \revision{float} \\
$H({p}_{aj=1}^{P-1})$ & Multi-layer perceptron with $P-1$ auxiliary label probabilistic encodings as inputs $p_{aj}$ & \revision{neural network function} \\
$x_i$ & word embedding of sentence $s_i$ & \revision{vector} \\ 
 & \revision{Output of variable after probability encoding} & \revision{float} \\ \hline
\textbf{Notation} & \textbf{Explanation} & \textbf{\revision{Source}} \\ \hline
$TP$ & The \revision{meta-feature} Throughput & \revision{Amazon Mechanical Turk} \\ 
$WT$ & The \revision{meta-feature} Worktime & \revision{Amazon Mechanical Turk} \\ 
$PC$ & The \revision{meta-feature} PC Agreement & \revision{Calculated}  \\ 
$TL$ & The \revision{meta-feature} Text Length & \revision{Calculated} \\ 
$SP$ & Combination options of the \revision{meta-feature} & \revision{Calculated} \\ 
\hline
\end{tabular}
\end{center}
\end{table}

\begin{figure}[!ht]
 \centering
 \includegraphics[scale=0.4]{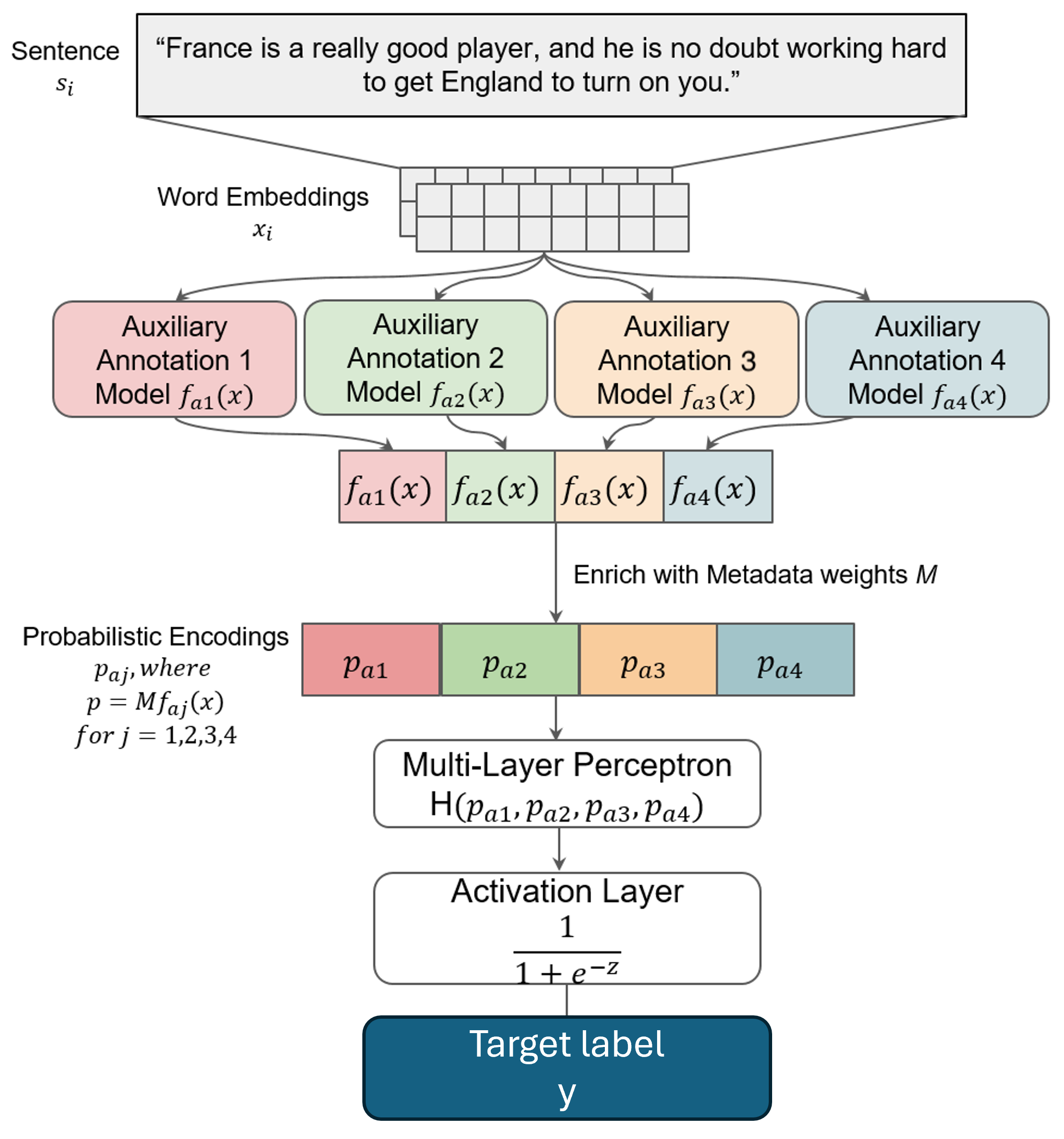}
 \caption{The proposed Metadata-Sensitive Weighted-Encoding Ensemble Model. The embeddings correspond to those generated through evaluating different classifiers and pre-trained tokenizers.}
 \label{fig:pipeline}
\end{figure}

\subsubsection{Example walkthrough}
In~\autoref{fig:pipeline}, we present our ensemble model that strategically incorporates meta-features from annotators to predict user behavior with enhanced accuracy. This approach moves beyond simplistic majority voting by factoring in behavioral meta-features such as annotator speed and agreement, which are crucial for a nuanced understanding of data quality. For illustration, consider a sentence annotated for gameplay strategies by four annotators (Figure~\ref{fig:pipeline}). Rather than relying solely on a majority vote to determine the target variable, we consider behavioral factors affecting annotation relevance.

In the first step, we model the target variable (for example, `deception' from~\autoref{tab:diplomacyannotation} or `popularity' from~\autoref{tab:popularityannotation}) as a function of its auxiliary variables. The ensemble model harnesses the auxiliary variables' collective predictive power to improve the target variable's predictive accuracy—an outcome critical for datasets with noisy, human-annotated inputs~\cite{ganaie2022ensemble}.

Subsequently, individual predictions for auxiliary labels are made. Each reflects a binary classification problem where the outcome indicates the presence or absence of specific conversational features. The classifier of each label outputs a posterior probability distribution of each label. We weigh each auxiliary label prediction in proportion to its prior distribution to account for dataset imbalance among these labels. 

The final stage entails synthesizing auxiliary label predictions enriched with annotator meta-features. This step adjusts the influence of each prediction, tailoring it to the annotation quality and shifting the importance of each output prediction by the quality of the annotation. The four probability encodings produced by the individual models of the auxiliary annotations are constructed into an encoding vector as an input to the ensemble layer. This vector is then passed through an activation layer to predict the target variable. The outcome is a refined vector, serving as an input to a multi-layer perceptron, which computes the target variable's probability.

For example, softmax outputs like [0.7, 0.2, 0.6, 0.9] for the labels 'deception,' 'treaty,' 'diplomat,' and 'lied' are concatenated as a vector. Meta-features transform them into weighted vectors, e.g., [0.35, 0.06, 0.42, 0.54], accounting for annotator variation. The perceptron then analyzes them to predict the target label.

Finally, in the third step, we combine the auxiliary labels' predictions to predict the target variable.The walkthrough illustrates how our method retains the breadth of annotations while privileging higher-quality inputs. It offers the model a refined lens to view quality annotations and incorporates semantic and annotator attributes into its predictions.

\subsection{Meta-feature variants}
\label{sec:metafeaturevariants}
We collect meta-feature values from the AMT task to represent annotation behavior. We incorporate these meta-features through a reparameterization of the probability encoding layer of the ensemble model $\{p_{aj}\}$ with meta-feature weights $\{M_l\}$ to obtain $\{M_lp_{aj}\}$. That means, for the $jth$ auxiliary variable $a_j$, it is enriched with the $lth$ meta-feature $\{M_l\}$. This variable is then passed through the probability encoding layer to obtain $\{M_lp_{aj}\}$, which is an encoding of the presence of $a_j$ given the meta-feature $\{M_l\}$.  These variants are combinations of the Throughput, Work time and PC Agreement. Text Length was not included in the combinations of meta-features because it is determined by the research investigator who provides the data for annotation. As such, it is not a feature specific to annotator behavior or quality, but rather a predefined attribute of the data.

Table~\ref{tab:encodinglayerweight} describes the variants of the meta-features, namely:

\begin{itemize} [noitemsep]
 \item \textbf{Average ratio of Throughput (TP1), Work time (WT1), PC Agreement (PC1):} The ratio of the normalized average meta-feature value for one observation to the maximum overall average, e.g., $M_{TP1} = {\{\overline{TP}\}_l} / {\text{Max}\{\overline{TP}\}}$
 \item \textbf{Variance ratio of Throughput (TP2), Work time (WT2), PC Agreement (PC2):} The ratio of the variance in meta-feature values for this observation to the maximum overall variance. eg. $M_{TP2} = (\{\text{Var}({TP})\}_l) / (\text{Max}\{\text{Var}({TP}_l)\})$
 \item \textbf{Quadratic ratios of Throughput (TP3, TP4) variances:} The variance of throughput feature values is converted with a quadratic function $f(x)=ax^2+bx+c$. TP3 is the negative $c<0$, and TP4 is the positive quadratic ratio $c>0$, calculated by taking a ratio against the maximum of this value over all input texts. 
 \item \textbf{Combination options (SP1, SP2, SP3):} An average of the sum of TP1 and TP2 for SP1, similarly for WT1 and WT2 for SP2, and PC1 and PC2 for SP3 is used as weights in the probabilistic encoding. $M_{SP1} = (M_{TP1} + M_{TP2}) / 2$
\end{itemize}

\begin{table*}[!h]
\caption{The variations of weighting the Encoding Layer (labeled ``Probabilistic Encodings'' in Figure 1). ${a_j}_{1}^4$ are the encodings for the game move, reasoning, share information, and rapport models.} 
\label{tab:encodinglayerweight}
\resizebox{0.8\textwidth}{!}{%
\begin{tabular}{lr}
\hline
\textbf{Metadata} & \textbf{Weighting} \\ \hline
\textbf{Throughput} &  \\ 
TP1: average  $\overline{TP}$ & (TP1*a$_{1}$, TP1*a$_{2}$, TP1*a$_{3}$, TP1*a$_{4}$) \\ 
TP2: linear variance V(TP) & (TP2*a$_{1}$, TP2*a$_{2}$, TP2*a$_{3}$, TP2*a$_{4}$) \\
TP3: Negative quadratic variance & (TP3*a$_{1}$, TP3*a$_{2}$, TP3*a$_{3}$, TP3*a$_{4}$) \\
TP4: Quadratic variance & (TP4*a$_{1}$, TP4*a$_{2}$, TP4*a$_{3}$, TP4*a$_{4}$) \\
 \hline
\textbf{Worktime} & \\
WT1: average $\overline{WT}$ & (WT1*a$_{1}$, WT1*a$_{2}$, WT1*a$_{3}$, WT1*a$_{4}$) \\
WT2: linear variance V(WT) & (WT2*a$_{1}$, WT2*a$_{2}$, WT2*a$_{3}$, WT2*a$_{4}$) \\
\hline
\textbf{Percentage Agreement} & \\
PC1: average $\overline{PC}$ & (PC1*a$_{1}$, PC1*a$_{2}$, PC1*a$_{3}$, PC1*a$_{4}$) \\
PC2: linear variance V(PC) & (PC2*a$_{1}$, PC2*a$_{2}$, PC2*a$_{3}$, PC2*a$_{4}$) \\
PC3: original PC per model & (PC$_a$*a$_{1}$, PC$_b$*a$_{2}$, PC$_c$*a$_{3}$, PC$_d$*a$_{4}$) \\
\hline
\textbf{Text Length} & \\
TL1: normalised number of characters $\frac{chars}{max_{chars}}$ & (TL1*a$_{1}$, TL1*a$_{2}$, TL1*a$_{3}$, TL1*a$_{4}$)\\ 
TL2: normalised number of words $\frac{words}{max_{words}}$ & (TL2*a$_{1}$, TL2*a$_{2}$, TL2*a$_{3}$, TL2*a$_{4}$)\\
\hline
\textbf{Combination Options} & \\
SP1: 0.5*(TP1+TP2) & (SP1*a$_{1}$, SP1*a$_{2}$, SP1*a$_{3}$, SP1*a$_{4}$)\\
SP2: 0.5*(WT1+WT2) & (SP2*a$_{1}$, SP2*a$_{2}$, SP2*a$_{3}$, SP2*a$_{4}$)\\
SP3: 0.5*(PC1+PC2) & (SP3*a$_{1}$, SP3*a$_{2}$, SP3*a$_{3}$, SP3*a$_{4}$)\\
\hline
\end{tabular}
}
\end{table*}


\subsection{Generating the prediction of the target variable}
We want to leverage the relationship between auxiliary labels and data quality to predict the target variable. For each meta-feature variant, we performed the following operations per annotation label. For each annotation label $j\in[1,P]$, we concatenated the softmax outputs $\{M_l\}\{p_{aj}\}$ of the individual models\footnote{The details of the individual model construction
are provided in Section 5.2.} trained to predict the annotation label into a 1xP-dimensional vector. The last layer of each individual model is a softmax layer, and the softmax outputs are being concatenated. After encoding this layer into a vector and enriching it with annotator information, we trained a three-layer perceptron model to predict the target variable in a final prediction step. 

\section{Experimental and Evaluation Setup}
\label{sec:expt}
Model development was based on the CLAff-Diplomacy dataset, described in Section 5.1, and validated on the CLAff-OffMyChest dataset, described in Section 5.2. Firstly, word embeddings were generated using the training set. The word embeddings are used as the input to train the individual auxiliary annotation models separately. The details of the individual model construction are described in Section~\ref{sec:indivmodels}. The softmax output of each annotation model is then appended into a probability encoding vector of length four. Subsequently, this vector will be enriched with the meta-features as weights representing the importance of each annotation. This vector is an input to train a multi-layer perceptron with a final softmax layer, which outputs the final annotation label. Experiments for MSWEEM were repeated at least twice each to ensure results were consistent between distinct instances. Finally, we report the average of the results.

We systematically removed each feature from the model for ablation analysis to understand its impact on the overall performance. For example, we might remove the 'deception' annotation and its corresponding meta-feature to see how it affects the final annotation label. We tested our model on different datasets for generalizability analysis to see how well it performs across various contexts. This helps us understand if our model is overfitting to the training data or if it can be applied to other similar tasks effectively.

\subsection{Training individual classifiers}
\label{sec:indivmodels}
The best-performing individual model architecture was used to construct auxiliary annotation models. They were selected by measuring the best macro F1 scores on a held-out test set. They were identified by testing two groups of models: non-transformer and transformer-based models. 

The datasets presented a huge class imbalance in terms of the annotated labels. We attempted to resolve this class imbalance issue in three ways. (1) We performed stratified sampling splits to partition our train-validation-test datasets to maintain the proportions of each class in the portion of the dataset. We used an 80-10-10 train-validation-test dataset split with stratified sampling for all experiments. That is, we performed training on 80\% of the training dataset. Hyperparameter tuning was done on a 10\% held-out validation set, and testing was performed on a 10\% test set.
(2) We tuned our loss functions during model training with the \textit{class\_weights} parameter in all experiments to factor the prior class distribution into the loss function.
(3) Finally, we used a macro F1 feature to assign a higher misclassification penalty on the minority class predictions instead of the accuracy measurement, which gives an inflated metric.

\subsubsection{Non-transformer models}
In the \textbf{first} group of non-transformer-based models, we used the Keras Tokenizer provided by the Tensorflow package to preprocess the input sentence $\{s_i\}$, filtering out punctuation and special characters before constructing a 220-long word embedding $\{x_i\}$ that is fed as input to the models. For all non-transformer-based models, we perform training using a batch size of 128 with an RMSprop optimizer and a learning rate of 1e-3 for 50 epochs.

These non-transformer-based models include:
\begin{itemize}[noitemsep]
    \item \textbf{CNN} \cite{kim-2014-convolutional} applies a sliding convolution window to the sequence before using a pooling operation to combine the vectors resulting from the different convolution windows. A one-layer 1D CNN with max-pooling is used over the input text.
    \item \textbf{LSTM} \cite{hochreiter1997long}, or Long Short-Term Memory networks, learns the sequence of an input text through one forward network. A one-layer LSTM with 64 hidden units is used.
    \item \textbf{LSTM-Attention} enhances the LSTM model with an Attention layer \cite{46201}, which emphasizes the embeddings of certain words in an input sentence while creating the vector. In addition, a mean attention layer is added to the LSTM outputs.
    \item \textbf{BiLSTM} \cite{liu-etal-2015-fine}, or bidirectional LSTM, is a sequence model that contains two LSTMs, one forward, one backward, to increase the context of a sequence given to the algorithm. A one-layer BiLSTM with 64 hidden units is used.
\end{itemize}

\subsubsection{Transformer-based models}
The \textbf{second} group of models are transformer-based models implemented with the SimpleTransformers python library\footnote{\url{https://github.com/ThilinaRajapakse/simpletransformers/}}. The input text is tokenized. Using different encoders, we generated the word embeddings corresponding to each model's pre-trained tokenizers. The models are trained with the recommended settings by the library: A batch size of 64 and a learning rate of 5e-5 for BERT and a batch size of 16 and a learning rate of 3e-5 for all the other models.
All models were trained for up to ten epochs with an early stopping criterion of 0.001 to reduce model overfitting. The model stops the training process when the L2-regularization loss between epochs changes to less than 0.001. We evaluated the following transformer-based models:

\begin{itemize}[noitemsep]
    \item \textbf{BERT} \cite{devlin2018bert}, or Bidirectional Encoder Representations from Transformers, pre-trained language representations on large text corpus like Wikipedia for unsupervised bidirectional models. 
    \item \textbf{Distilbert} \cite{sanh2019distilbert} leverages knowledge distillation during the model pre-training phase to reduce the size of the BERT model.
    \item \textbf{XLNet} \cite{DBLP:journals/corr/abs-1906-08237} is an auto-regressive language model which calculates the joint probability of a sequence of word tokens based on a recurrent transformer architecture.
    \item \textbf{RoBERTa} \cite{DBLP:journals/corr/abs-1907-11692} optimizes BERT by modifying its key hyperparameters, removing next-sentence pre-training objective and performs training with larger mini-batches and learning rates.
\end{itemize}

A consistent input vector size of length 220 is used for all the non-transformer-based models. A consistent input vector with a length of 128 is used for all the transformer-based models. In both classifiers, the space complexity is O(1) because the length of the input text does not change the size of the input vector and thus does not affect the storage requirements for the model~\cite{tsironi2017analysis}. The time complexity of all models is O(n), where n is the number of convolutional layers present in the model~\cite{shah2022time}.

\subsubsection{Baselines}
\label{subsec:baselines}
We applied two approaches from previous work as baselines to test their effectiveness in improving classification performance. 
First, we tested \textbf{MACE}~\cite{paun2018comparing,hovy-etal-2013-learning}, or Multi Annotator Competence Estimate, which uses a generative Bayesian probability method to model annotator credibility. It first characterizes prior parameters representing the behavior of annotators through a Dirichlet distribution and then parameterizes each annotator's behavior through a beta distribution. A posterior Bernoulli distribution then characterizes every annotation data point. This distribution serves as annotator metadata in the MACE setup. Finally, it predicts a binary label. We adapted the Stan code released by the authors and trained the model using our datasets for our experiments. This baseline uses a probability-based approach. It assumes meta-feature independence, as compared to our neural network approach which accounts for complex relationships between features ~\citep{10.48550/arxiv.1909.12911}.

Second, we tested a previous approach by~\citet{geva2019we}, which showed that models perform better when exposed to annotator IDs. The models can differentiate annotator styles and capture nuances of annotators as meta-features. In our implementation, we concatenate annotator IDs as a meta-feature to a pre-trained BERT model as an additional feature. We used a pre-trained BERT-base model \cite{devlin2018bert} with the same parameters as the original work: fine-tuning for three epochs, batch size 10, and learning rate of $2 \times 10^{-5}$. This baseline distills annotation features by comparing vectors with the same annotator IDs. Our method directly appends the meta-features from the annotators. We compare our method with this baseline to learn whether it is better to directly capture annotator meta-features rather than distilling them.

\subsection{Evaluation}

Initially, we train various model variations specifically for auxiliary text classification tasks within the CLAff-Diplomacy dataset. Next, the top-performing model variant is chosen for each task, and its probabilistic encoding layer is employed to train ensemble models that incorporate diverse types of metadata. We assess the improvement achieved by incorporating these additional metadata variants compared to using the individual model variants alone (referred to as base models in~\autoref{tab:bestofpipelinefull}).

Text classification approaches that rely on neural networks are often black boxes where feature importance is hard to diagnose. Therefore, we increased their interpretability and established meta-feature contributions in the MSWEEM under different limiting conditions. We validated both the ensemble approach and the importance of metadata in three main ways:
\begin{itemize}
    \item We compared the predictive performance against an ensemble-based model without meta-features (referred to as base models in Table~\ref{tab:bestofpipelinefull}). We also compared the predictive performance of our approach to baseline models from literature, as described in~\autoref{subsec:baselines}, that also incorporate annotator metadata in their architecture.
\item We conducted an \textbf{ablation analysis} of feature importance vis-a-vis dataset size. To do so, we systematically reduced the training data size and noted its effect on predictive performance. 
\item We also established the \textbf{generalizability} of MSWEEM to a different dataset and knowledge hierarchy, i.e., the CLAff-OffMyChest dataset.
\item Finally, we examined the \textbf{generalizability} of the relationship of one annotator cohort's performance with a different one. In this case, we compared workers with and without the Master's qualification on the AMT platform.
\end{itemize}

\section{Datasets}
\label{sec:datasets}
The distinction between behavioral and semantic labels is crucial in our choice of datasets. Behavioral labels serve as our target variable and are derived from the actions or reactions of individuals. They depend on users' interpretation and interaction. On the other hand, semantic labels are our auxiliary variables, and they are derived directly from the content's inherent attributes.

\begin{table}[!ht]
\caption{CLAff-Diplomacy final label and auxiliary label information (acquired label is Deception).}
\label{tab:diplomacyannotation}
\begin{center}
\small
 \resizebox{1.0\textwidth}{!}{
\begin{tabular}{p{3cm}p{4cm}p{7cm}}
\hline
\textbf{Label} & \textbf{Elaboration} & \textbf{Example \newline Messages} \\ \hline
Deception & Whether or not the receiver thinks the speaker is lying & ``I think I can top Germanys offer if youre interested." \newline
``I want it to be clear to you that youre the ally I want."\\
\hline
Game Move & Actual or suggested game move & ``I am committed to supporting Munich holding" \newline ``Make sure you don't move Munich so that it can take my support." \\ \hline
Reasoning & Justify a move, guess what moves might happen next, or discuss a move that already happened & ``If you took Marseilles, I would be stronger against England" \newline ``France is a really good player, and he is no doubt working hard to get England to turn on you."\\ \hline
Share Information & Share information related to other players & ``France is a really good player, and he is no doubt working hard to get England to turn on you." \newline ``And anything that's bad for Russia right now is good for Austria."\\ \hline
Rapport & Build rapport through compliments, concerns, reassurances or apologies & ``I won't hold it against you" \newline ``I'm going to keep helping you as much as I can." \\ \hline
\end{tabular}
}
\end{center}
\end{table}

We obtained the annotation-level meta-features of two public, conversational text datasets to develop and validate MSWEEM, which have been published as a part of the AAAI Affective Content Analysis Workshop's CLAff Shared Tasks~\cite{DBLP:conf/aaai/JaidkaSLCU20,jaidkaeditorial}. The training and testing dataset was the CLAff-Diplomacy dataset with 11,366 sentences with four auxiliary labels and target variable \textit{$y$=deception}. The validation dataset was the CLAff-OffMyChest dataset with 17,473 Reddit posts and comments with four auxiliary labels and target variable \textit{$y$=popularity}.\footnote{Training dataset: The CLAff-Diplomacy dataset at https://sites.google.com/view/affcon2021/CLAff-shared-task \\ Validation dataset: The CLAff-OffMyChest dataset at https://sites.google.com/view/affcon2020/CLAff-shared-task}~\cite{DBLP:conf/aaai/JaidkaSLCU20}. 
\revision{The meta-features Throughput, Worktime were obtained from the AMT platform. These features are available when one posts tasks on the platform. We sought these data from the original authors of the datasets. We calculated the PC Agreement and Text Length meta-features.}
The class distributions of the data (the target and auxiliary labels) are provided in Figure \ref{fig:datasetsplit}. 

\begin{figure}[!ht]
 \centering
 \includegraphics[width=0.7\textwidth]{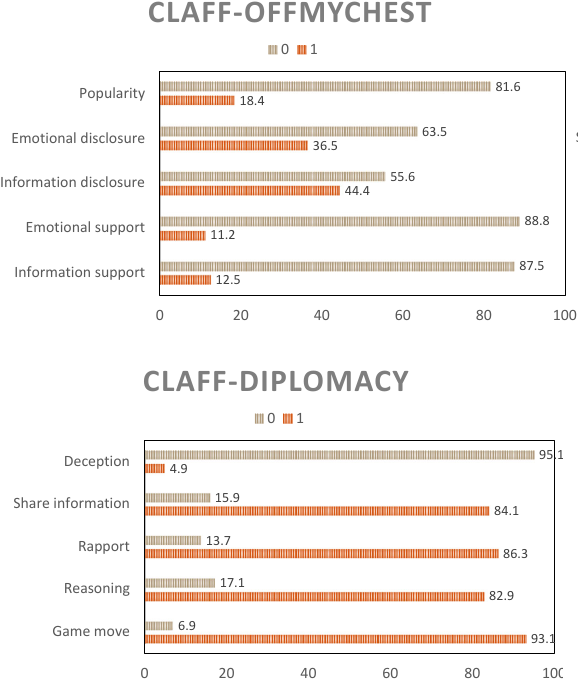} 
\caption{The label distributions for the CLAff-OffMyChest and CLAff-Diplomary datasets. Because both datasets present very imbalanced class distributions, we adjusted for the skew by tuning the loss functions during the model training. 1 means the presence of the class, while 0 means the absence of the class.}
\label{fig:datasetsplit}
\end{figure}

\subsection{CLAff-Diplomacy Dataset}
The CLAff-Diplomacy dataset was released as a part of the CLAff Shared Task 2021~\cite{jaidkaeditorial}\footnote{https://github.com/kj2013/claff-diplomacy} and comprises chat messages exchanged between players in
an online game called Diplomacy. The text classification problem involved better understanding the rhetorical strategies used by the players to negotiate support from each other in pairwise conversations. The dataset was first introduced by~\citet{peskov-etal-2020-takes}, which included utterance-level annotations about whether the receiver trusted the speaker. This constitutes the target variable for our modeling problem, denoted by the receiver's anticipation of $y$=Deception. The dataset was extended by~\cite{jaidkaeditorial,jaidka2024takes} to identify whether the players were discussing game moves, reasoning their strategy, sharing information, or building rapport ($a$=\{Game Move, Reasoning, Share Information, and Rapport\}).

In the context of the CLAff-Diplomacy dataset, ``Deception" serves as a behavioral label because it captures an aspect of the players' interaction strategies—specifically, whether the receiver perceives the sender's message as deceitful or not. This relates directly to the players' behavior in the game and is determined by the psychological state and intention of the sender as interpreted by the receiver. It is an assessment of the sender's actions (the act of deceiving) rather than the content of the conversation itself. On the other hand, the other labels, such as ``Game Move," ``Reasoning," ``Share Information," and ``Rapport," are semantic because they relate to the content and subject matter of the conversation. These labels classify the nature of the information being exchanged (e.g., discussing game tactics, providing reasons for a particular strategy) or the type of social interaction occurring (e.g., building a connection or rapport), which are based on the textual content rather than the behavior of the interlocutor.

Finally, in the CLAff-Diplomacy dataset, each annotator's behavioral signals of throughput $TP$ and worktime $WT$ were available. Traditional annotation quality metrics comprised overall percentage agreement and input text length meta-feature. The average inter-annotator agreement was 80.8\%.~\autoref{tab:diplomacyannotation} provides label information and example observations.

\begin{table}[!ht]
\caption{CLAff-OffMyChest final label and auxiliary label information (acquired label is Popularity).}
\label{tab:popularityannotation}
\begin{center}
\small
 \resizebox{1.0\textwidth}{!}{
\begin{tabular}{p{3cm}p{4cm}p{7cm}}
\hline
\textbf{Label} & \textbf{Elaboration} & \textbf{Example \newline Messages} \\ \hline
Popularity & Whether a comment's upvotes are greater than or equal to the overall average number of upvotes & ``I don't know who is telling you this is bad parenting, but they are just flat out wrong." \newline
``I encourage you to report it, (...) I know how awful it can be to be raped (...)"\\
\hline
Emotional disclosure & Mentions the author’s feelings. & ``I love her and all, but I can leave her." \newline ``You're the reason that me has died?" \\ \hline
Information disclosure & Contains at least some personal
information about the author. & ``Hey, I don't do the exclusive partner thing." \newline ``(Extra info: I was also constantly in trouble for hurting her feelings, at times I would be called a bitch."\\ \hline
Emotional support & Offers sympathy, caring, or encouragement & ``Cry as much and as long as you need." \newline ``I'm 100\% with you that the OP should not pursue sex."\\ \hline
Information support & Offers specific information,
practical advice, or suggesting a course of action & ``	
6 - Hire a CPA with a good reputation!" \newline ``Ask for it to stay with you for every morning. :)  I know how you feel." \\ \hline
\end{tabular}
}
\end{center}
\end{table}

\subsection{CLAff-OffMyChest Dataset}
The CLAff-OffMyChest dataset~\cite{DBLP:conf/aaai/JaidkaSLCU20} comprises sentences from comments posted to two Reddit communities, /r/OffMyChest and /r/CasualConversation. These comments were made in response to original posts discussing families and relationships, filtered for keywords such as \textit{boyfriend, girlfriend, husband, wife, gf, bf}. In the CLAff-OffMyChest dataset, ``Popularity'' is a behavioral label that measures the reaction of the community to a comment—captured quantitatively as the number of upvotes—which is a direct outcome of user behavior in response to the comment. It reflects the communal behavior in valuing the content of the comment, which is contingent on the community's social dynamics and emotional resonance. Therefore, the target variable is $y$=Popularity, transformed into a 0/1 class from the raw number of upvotes received for the comment. A 1 value reflects that the comment was more popular than average (received more upvotes than an average comment). 

Each observation in the dataset was annotated for features of emotional/information disclosure/support  $a$=\{Emotional disclosure, Information disclosure, Emotional support, Information support\}. These labels are semantic because they categorize the type of disclosure or support expressed in the comments. These labels concern the nature of the information or emotion shared in the comment, indicative of the content's characteristics rather than the community's behavior in response to it.

Two sets of meta-features are available for each observation, corresponding to the disclosure and support labels, respectively. The average inter-annotator agreement of the labeled dataset was 77.9\%.~\autoref{tab:popularityannotation} reports label information and example observations.

\subsection{Data Preparation}
The annotations and metadata information about both datasets, CLAff-Diplomacy and CLAff-OffMyChest, were sourced from their AMT annotations. Each data point had the corresponding annotator metadata describing the annotator's throughput and worktime.
We calculated the percentage agreement for each assigned label from the aggregated annotations. We separately calculated text length for the corresponding annotated texts.

Each dataset has one target variable and four auxiliary labels. It also has meta-features related to each observation in the dataset. A pairwise correlation analysis between the labels and the meta-features (reported in the Supplementary Materials) shows that the meta-features are non-collinear and capture different aspects of annotation attention checks. For the CLAff-Diplomacy dataset, a higher percentage agreement is associated with lower throughput (r = -0.44, p $<$ 0.001) and greater work time (r = 0.04, p $<$ 0.001). 

Dataset preprocessing was performed to obtain consolidated behavioral meta-features as the average and variance of individual values. The feature distributions for CLAff-Diplomacy are provided in~\autoref{fig:diplomacymetadataplots}. The throughput meta-feature is markedly skewed compared to work time and text length. The percentage agreement feature has values across the normalized range of 0 and 1. The feature distributions for CLAff-OffMyChest are in the supplementary materials. 

\begin{figure*}[!ht]
 \centering
 \includegraphics[width=\textwidth]{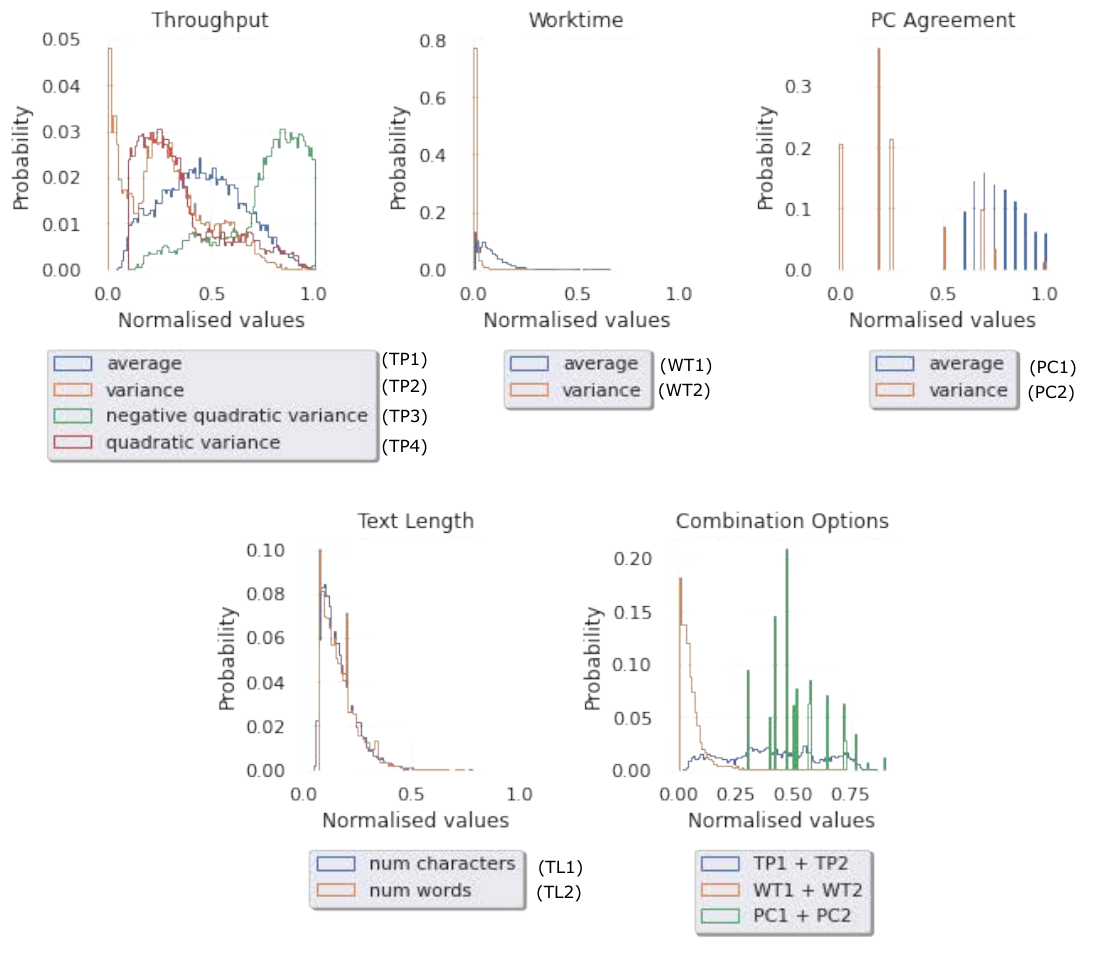} 
 \caption{The probability distributions of the different meta-feature variants for the CLAff-Diplomacy dataset. The most informative features are expected to have a wider spread of values.}
 \label{fig:diplomacymetadataplots}
\end{figure*}


\begin{table*}[ht]
\caption{\small Individual F1 scores for the (a) CLAff-Diplomacy and (b) CLAff-OffMyChest datasets. The best performing model in each column is used as input in the ensemble approach. We also represent F1 scores from previous work which performed predictions using weighting scheme based on annotator information.}
\vspace{-\baselineskip}
\label{tab:individualscoresdataset1}
\begin{center}
 \resizebox{1.0\textwidth}{!}{
\begin{tabular}{lrrrrr}
\hline
\textbf{Model} & \multicolumn{5}{c}{\textbf{F1 Score}} \\ \hline

\multicolumn{6}{c}{\textbf{CLAff-Diplomacy dataset  (N = 11,366)}} \\ \hline
& \textbf{Game Move} & \textbf{Reasoning} & \textbf{Share \newline Information} & \textbf{Rapport}\\ \hline
CNN & 0.404 &\textbf{ 0.518} & 0.485 & 0.469 \\
LSTM &\textbf{ 0.481} & 0.515 &\textbf{ 0.502} & 0.459  \\
LSTM-Attention & 0.484 & 0.486 & 0.489 & \textbf{0.475} \\
BiLSTM & 0.480 & 0.483 & 0.480 & 0.456 \\
BERT & 0.484 & 0.451 & 0.458 & 0.466  \\ 
Distilbert & 0.484 & 0.451 & 0.458 & 0.466  \\
XLNet & 0.484 & 0.451 & 0.458 & 0.466 \\
RoBERTa & 0.484 & 0.451 & 0.458 & 0.466  \\ \hline
\textbf{Baselines with annotator metadata} & & & & & \\
\hline
MACE (\cite{paun2018comparing})  & 0.501 & 0.615 & 0.600 & 0.582 \\ 
Geva et al.\cite{geva2019we} & 0.513 & 0.485 & 0.513 & 0.474\\ 
\hline
\multicolumn{6}{c}{\textbf{(b) CLAff-OffMyChest dataset  (N = 17,473)}} \\ \hline
\textbf{Model} & \multicolumn{5}{c}{\textbf{F1 Score}} \\ \hline
& \textbf{Emotional Disclosure} & \textbf{Information Disclosure} & \textbf{Emotional Support} & \textbf{Information Support} \\
CNN & 0.642 & 0.651 &\textbf{ 0.710} & 0.664 \\ 
LSTM & 0.665 & 0.671 & 0.709 &\textbf{ 0.668} \\ 
LSTM-Attention & \textbf{0.670} & \textbf{0.672} & \textbf{0.742 }& 0.633  \\ 
BiLSTM & 0.658 & 0.666 & 0.665 & 0.652  \\
\hline
\textbf{Baselines with annotator metadata} & & & & & \\
\hline
MACE \cite{paun2018comparing} & 0.763 & 0.831 & 0.645 & 0.667 \\ 
Geva et al. \cite{geva2019we} & 0.683 & 0.669 & 0.781 & 0.690 \\ 
\hline
\end{tabular}
}
\end{center}
\end{table*}
\section{Results and Validation}
\label{sec:results}
\subsection{Training individual classifiers}
\label{sec:indiv_classifiers}
In Table \ref{tab:individualscoresdataset1}, we report our experiments with different model variations for the auxiliary text classification tasks in the CLAff-Diplomacy dataset (trained \textbf{without} meta-features). The four auxiliary tasks are to predict game moves, reasoning, sharing information, and rapport. We calculated the average F1 score of each model to select the best-performing individual model. For the CLAff-Diplomacy dataset, the LSTM model had the best performance on predicting game moves (F1=0.481) and sharing information (F1=0.502), while CNN had better performance for reasoning (F1=0.518) and rapport (F1=0.475). For the CLAff-OffMyChest dataset, the LSTM-Attention model performed best with emotional disclosure, information disclosure, and emotional support. The LSTM model performed the best in information support. For the final architecture, we chose the set of models for the auxiliary variables from~\autoref{tab:individualscoresdataset1} based on the F1 scores.

\begin{table}[!ht]
\caption{F1 Scores for best performing MSWEEM variants for the CLAff-Diplomacy dataset. This is the prediction of the entire ensemble model that aggregates the auxiliary predictions weighted by annotator metadata. Base pipeline represents no weightage using AMT metadata features. The probabilistic encoding are joint using a Multi-Layer Perceptron.}
\label{tab:bestofpipelinefull}
\begin{center}
\begin{adjustbox}{totalheight=0.7\textheight}
 \resizebox{0.9\textwidth}{!}{%
\begin{tabular}{lr}
\hline
\textbf{Model} & \textbf{F1 Score} \\ \hline
Best direct prediction & 0.487 \\
\hline
\multicolumn{2}{l}{\textbf{CNN}} \\ 
Base  & 0.380 \\ 
TP2:  linear variance $V(TP)$ & 0.445 \\
WT1:  average $\overline{WT}$ &  0.487 \\
PC1:  average $\overline{PC}$  &\textbf{ 0.553 }\\
TL2:  number of words &  0.518 \\
SP1: $0.5*(TP1+TP2)$ & 0.433 \\
\hline
\multicolumn{2}{l}{\textbf{LSTM}} \\ 
Base &  0.456 \\ 
TP1:  average $\overline{TP}$ & 0.456 \\ 
WT1:  average $\overline{WT}$& 0.494 \\ 
PC1:  average $\overline{PC}$& \textbf{0.553} \\ 
TL2:  number of words & 0.463 \\ 
SP1: $0.5*(TP1+TP2)$  & 0.379\\
\hline
\multicolumn{2}{l}{\textbf{LSTM-Attention}} \\ 
Base &  0.340 \\ 
TP2:  linear variance $V(TP)$ & 0.444 \\ 
WT2:  linear variance $V(WT)$ & \textbf{ 0.487} \\ 
PC1: average $\overline{PC}$ & 0.432 \\
TL2: normalised number of words & 0.421 \\
SP2: $0.5*(WT1+WT2)$ &  0.487 \\
\hline
\multicolumn{2}{l}{\textbf{BiLSTM}} \\ 
Base & 0.326 \\ 
TP1:  average $\overline{TP}$ & 0.487 \\
WT1:  average $\overline{WT}$ & 0.329 \\
PC1:  average $\overline{PC}$ &\textbf{ 0.490} \\
TL1:  number of characters  & 0.476 \\
SP1: $0.5*(TP1+TP2)$  & 0.441 \\
\hline
\multicolumn{2}{l}{\textbf{BERT}} \\ 
Base  & 0.487 \\ 
TP1:  average $\overline{TP}$  & 0.466 \\ 
WT2:  linear variance $V(WT)$ & \textbf{0.487}\\
PC3:  original PC per model & 0.466 \\
TL1:  number of characters  & 0.047 \\ 
SP2: $0.5*(WT1+WT2)$  & 0.487 \\
\hline
\multicolumn{2}{l}{\textbf{Distilbert}} \\ 
Base & 0.487 \\ 
TP1:  average $\overline{TP}$ & 0.420 \\ 
WT1:  average $\overline{WT}$  &0.332 \\
PC3:  original PC per model  & 0.466 \\
TL2: number of words 1 & 0.421 \\
SP2: $0.5*(WT1+WT2)$ & \textbf{0.487} \\
\hline
\multicolumn{2}{l}{\textbf{XLNet}} \\ 
Base & \textbf{ 0.487}\\ 
TP1:  average $\overline{TP}$ & 0.487 \\ 
WT1:  average $\overline{WT}$ & 0.311 \\
PC1:  average $\overline{PC}$ & 0.418 \\
TL2: number of words & 0.487 \\
SP1: $0.5*(TP1+TP2)$ &  0.047 \\
\hline
\end{tabular}
}
\end{adjustbox}
\end{center}
\end{table}

\subsection{Ensemble models}
\label{sec:results2}
To construct the ensemble models, we first enriched the individual models with annotator meta-features and performed training and testing with these enhanced features. We then trained the ensemble models using the probabilistic encoding layer representing predictions from the best individual models. Each prediction was weighted with a variant of metadata. These scores now become the baselines for enriching meta-features. From the results, we selected the best-performing meta-features of each variant for each baseline model and have presented the average macro-F1 scores in~\autoref{fig:bestofpipelinefull}. Enriching the model with meta-features for all model architectures improved the prediction quality, lending weight to our hypothesis of improving predictive value in an ensemble setup with auxiliary variables. Finally, an extensive validation with ablated and unseen data is reported.

The results for the best-performing enriched ensemble models (highest F-1 score on the held-out validation dataset) are presented in the orange bars of Figure \ref{fig:bestofpipelinefull}. The full results are in Table~\ref{tab:bestofpipelinefull}. Table~\ref{tab:bestofpipelinefull} varies the types of weights to examine which type of meta-data is best for increasing the predictive capability of the target variable. The model enriched with annotator metadata performed better than the ``Base" model in the Table, representing the final model without any metadata combination. 

\begin{figure}[ht]
 \centering
 \includegraphics[width=.7\textwidth]{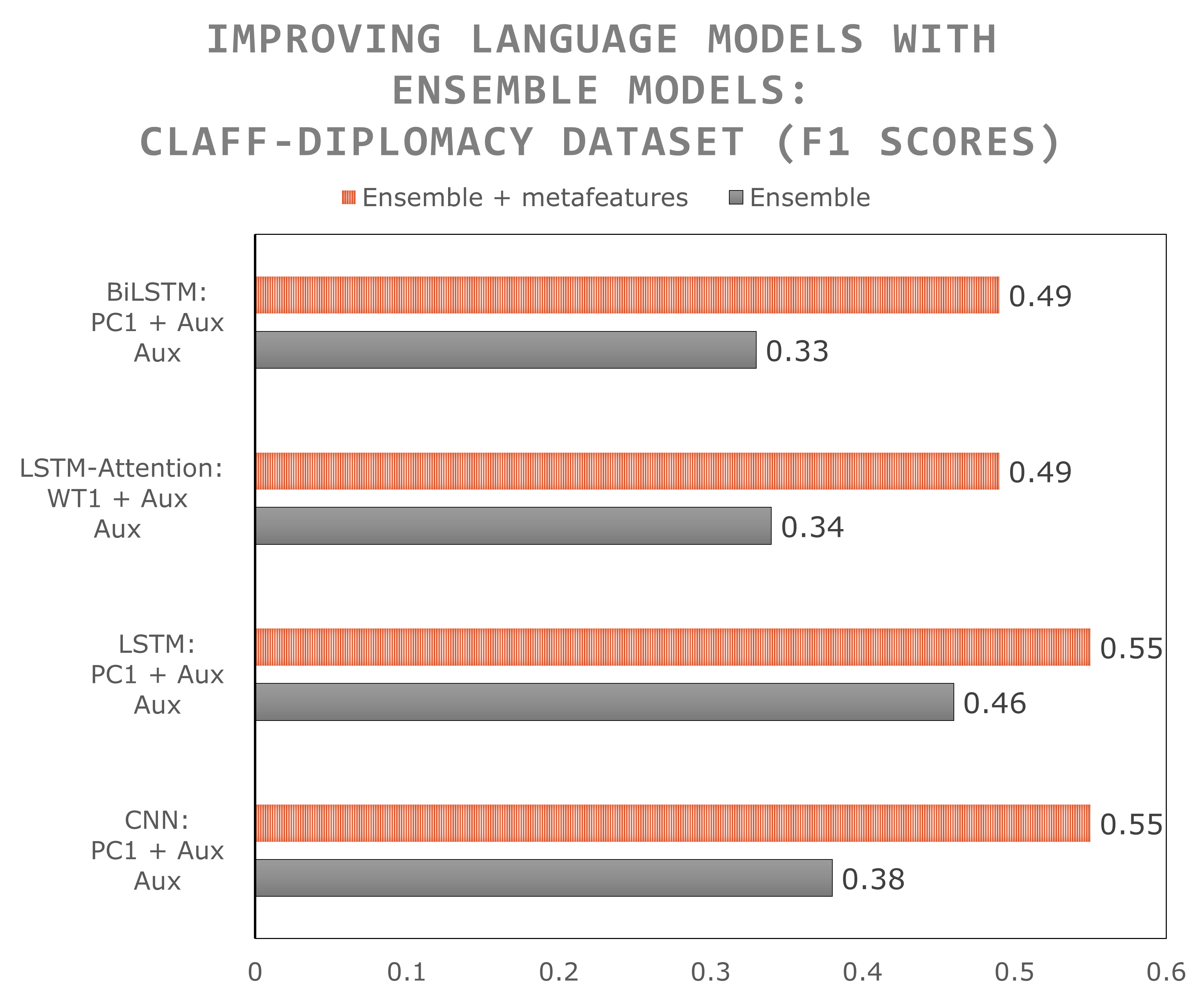} 
 \caption{Predictive performance on the CLAff-Diplomacy dataset considering the ensemble models in Table~\ref{tab:bestofpipelinefull}. Enriched ensemble models improve upon simple ensemble models (with only auxiliary attributes). PC1, WT1, and other meta-feature variants are described in Section~\ref{sec:metafeaturevariants} `Meta-feature variants' and in Figure~\ref{fig:diplomacymetadataplots}.}
 \label{fig:bestofpipelinefull}
\end{figure}

On the CLAff-Diplomacy dataset, base ensemble models outperform direct prediction with an average of 1\%, with the CNN and LSTM-based ensemble model leading with a performance improvement of 9\%. The model that shows the most significant improvement with AMT meta-features is CNN with percentage agreement (Base F1=0.380, with meta-features F1=0.553). On average, the models improve by 14\% when enhanced with meta-features (red bars vs. grey bars). Non-transformer models show notable improvements in an ensemble model combined with meta-features. Incorporating metadata has overcome their natural limitation of ignoring relative information and losses due to pooling. On the other hand, transformers use self-attentive mechanisms, where the ensemble approach plateaus at a high level of performance. 

\subsection{Dataset size ablation analysis}
To investigate the influence of dataset size on the ensemble model, we evaluated the predictive performance of the CNN-MSWEEM variants with differently sized stratified samples from the CLAff-Diplomacy dataset. These experiments are conducted with a dataset size beginning with 250 points to the maximum number of sentences. Figure~\ref{fig:randomcontrol} plots the F1-score as a linear function of the dataset size. The blue line depicts the performance trend of models trained without meta-feature weights as the control experiment. Their performance is nearly always the lowest, indicating that meta-feature enrichment improves predictive performance. The trend is similar for second and third-order plots. In general, meta-feature enrichment improved F1 scores by an average of 13\% across all dataset sizes. The steepest degradation is seen for the model with throughput information, which starts as the best but degrades quickly. The performance increment persists over different training set sizes. This suggests an optimal dataset size for model performance, regardless of the weights added, which occurs around 6000 data points. Such information could be used when constructing crowdsourcing surveys, where researchers could provision their resources and funding to collect around 6000 data points.

\begin{figure}[!ht]
\centering
 \includegraphics[width=.5\textwidth]{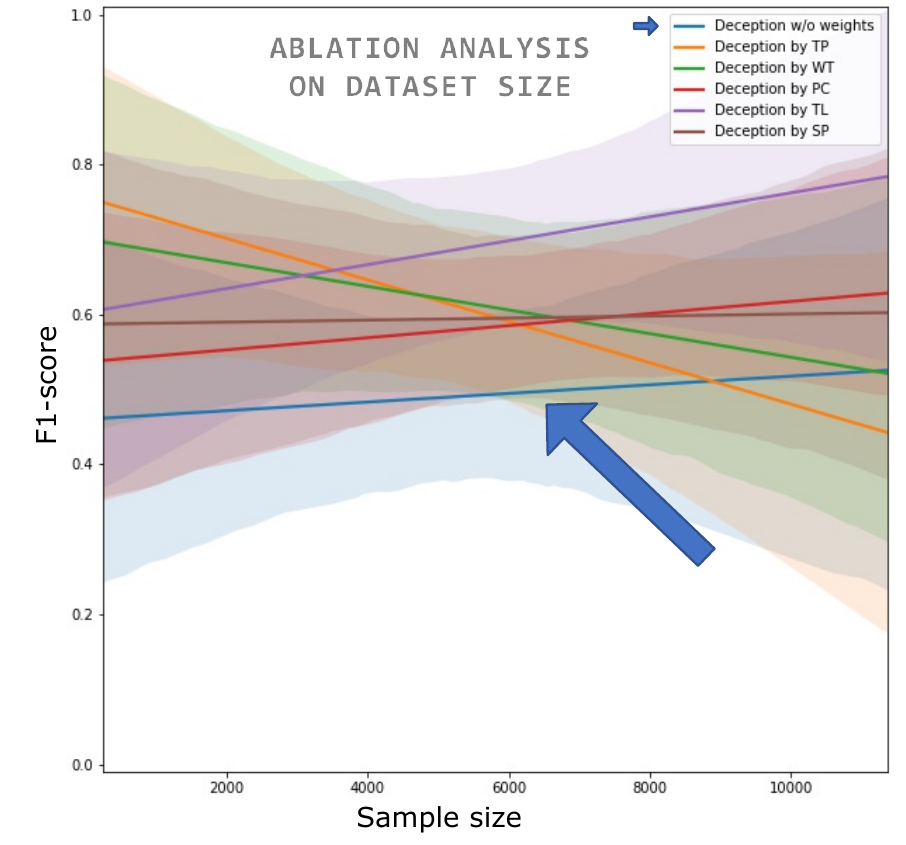}
\caption{Ablation analysis for the CLAff-Diplomacy dataset: Effect of dataset size on the predictive performance.}
\label{fig:randomcontrol}
\end{figure}

\subsection{Generalizability to the CLAff-OffMyChest dataset}
We validated MSWEEM on CLAff-OffMyChest, where the four auxiliary labels were emotional disclosure, information disclosure, emotional support, and information support. The target outcome was popularity. While the base ensemble model can perform worse than the direct prediction model, enhancing the model through meta-features generally improves the model performance. The results of the ensemble models are in~\autoref{fig:bestofpipeline_offmychest}. Meta-feature-enriched ensemble models (red bars) improved upon simple ensemble models (grey bars). This increase in performance is most prominent in the LSTM-Attention variant, where the positive quadratic variant of Work Time (F1=0.475) performed 19\% better than the base model (F1=0.285). On average, the addition of meta-features improves the model by 12\%. The CLAff-Diplomacy dataset shows the greatest improvement with the text length metadata (F1=0.285 to F1=0.449), but the CLAff-OffMyChest dataset showed the greatest improvement with the percentage agreement metadata.

\begin{figure}[!ht]
\centering
 \includegraphics[width=.70\textwidth]{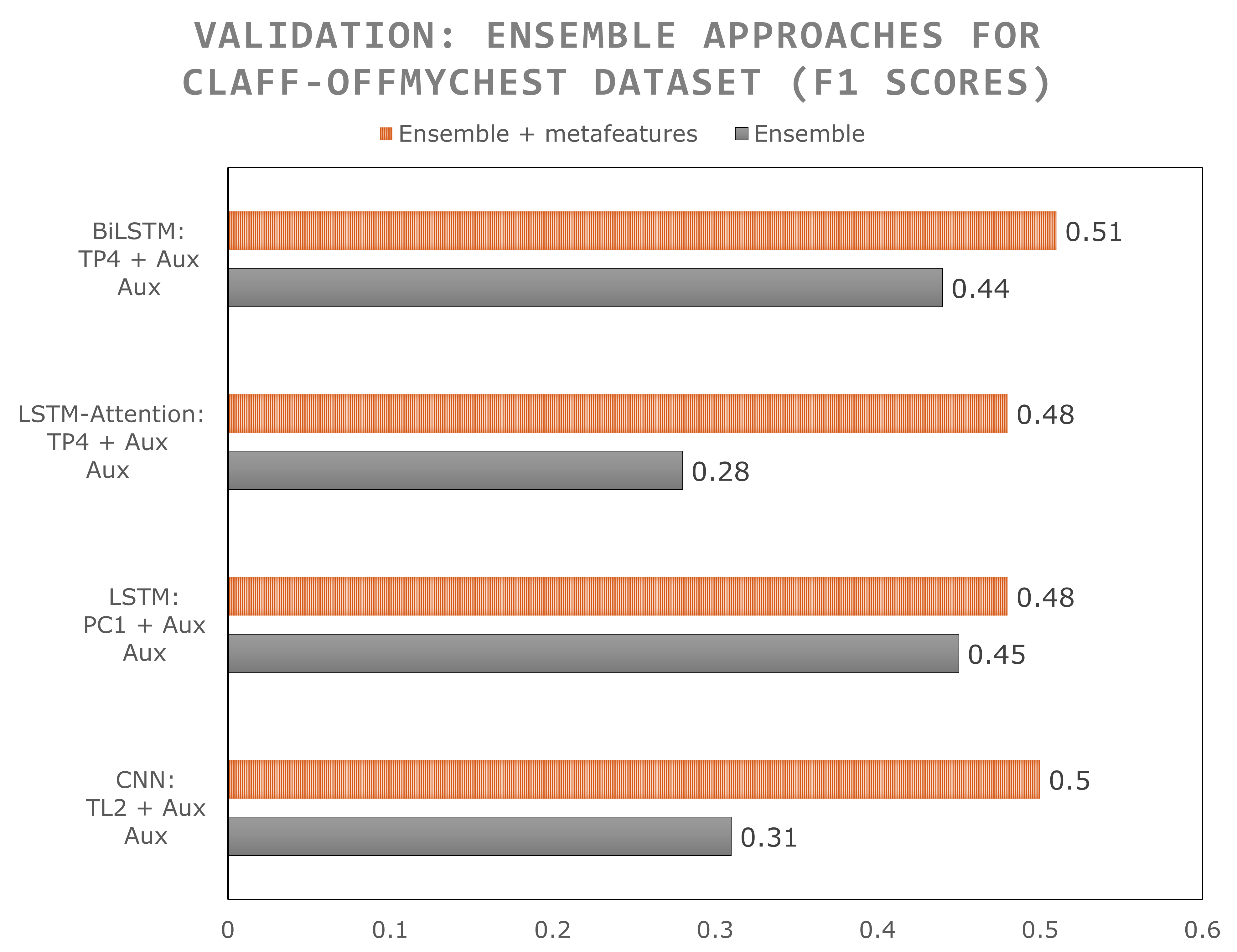} 
\caption{Predictive performance on the CLAff-OffMyChest dataset: Enriched ensemble approaches outperform simple ensemble models. }
\label{fig:bestofpipeline_offmychest}
\end{figure}
\subsection{Generalizability to different annotator cohorts}
We designed an additional task involving annotating the grammatical quality of short arguments about politics available from recent work~\cite{verma2024auditing}. Annotators were required to evaluate the grammatical quality of the post on a scale of 1 to 5, where 5 denoted perfect grammar. The dataset comprised 960 rows, and each row required five annotations. Our goal was to compare the inter-relationships between annotator behavior and performance across different cohorts, and we chose to compare AMT workers with and without Master's qualifications. First, we launched the task only for AMT workers with a Master's qualification, residing in the United States, who had completed at least 5000 HITs with a 95\% approval rating - standard criteria recommended by prior work~\cite{chandler2014nonnaivete}. The Master's qualification distinguishes a small cohort of workers who qualified for the label due partly to its bespoke work quality. In this manner, 4800 annotations were collected. Next, the task was launched for regular workers from the United States who had completed at least 5000 HITs with a 95\% approval rate.

Figure~\ref{fig:mastervsnormal} offers a comparative analysis of the annotator meta-features for this dataset, where all the metrics have been rescaled to a 0 to 1 range for ease of visualization. a compares the distribution of worktime for Master's and normal AMT annotators. The y-axis represents the density of observations; therefore, for any given worktime (rescaled to a 0 to 1 range for ease of visualization) on the x-axis, a higher value on the y-axis indicates a higher concentration of observations. The Master's density is also more sharply peaked, suggesting that Master's workers have more consistency around the peak worktimes. At the same time, there is wider variation in speeds, and hence a wider variation in performance, among ordinary workers. However, the data suggests that Master's workers may comprise two subgroups reflected in the two peaks in worktime distribution at which tasks were completed. There may be performance differences between these two groups. Figure~\ref{fig:mastervsnormal}b reports the comparative analyses of throughput and worktime between master and ordinary Amazon Mechanical Turk (AMT) annotators. It suggests that annotators with a Master's qualification are significantly faster (Mean time for Master's = 377.56 seconds vs Mean time = 1336 seconds; p = 2.2e$^{-16}$) and they push out more individual work (Mean throughput for Master's = 620 tasks vs Mean throughput for Normal workers = 10.4 tasks; p = 2.2e$^{-16}$)  than normal annotators. Finally, Figure~\ref{fig:mastervsnormal}c reports the association of the interactive effect between its association with the efficacy of annotators, calculated as the ratio of the difference between the annotator's rating and the mean rating for a data point, to the maximum difference over the entire dataset. It reports the fitted estimates from the logistic regression model of the annotators' alignment with the rating consensus, fitted as a function of the speed, throughput, and the moderation effect of the master's qualification on these factors. The model summary is reported in the supplementary materials. The regression analysis depicted in the second plot illustrates the significant impact that an increase in speed has on the decrease of the Master's annotators' alignment with the consensus ($\beta$ = -0.09, s.e. = 0.03, p = 0.01*), but not for normal annotators.

\begin{figure}[ht]
 \centering
 \includegraphics[width=.4\textwidth]{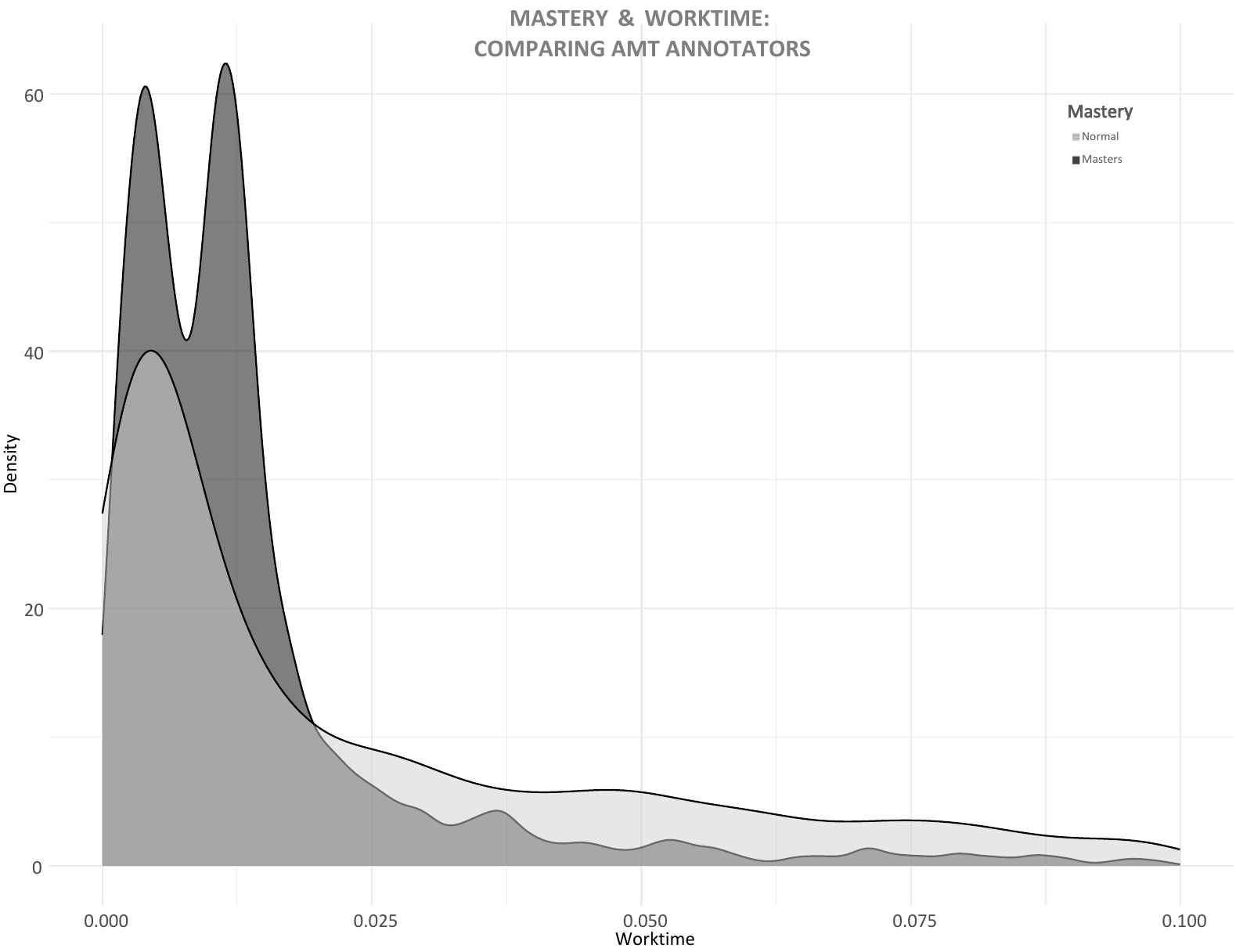}\\
 \includegraphics[height=.25\textwidth]{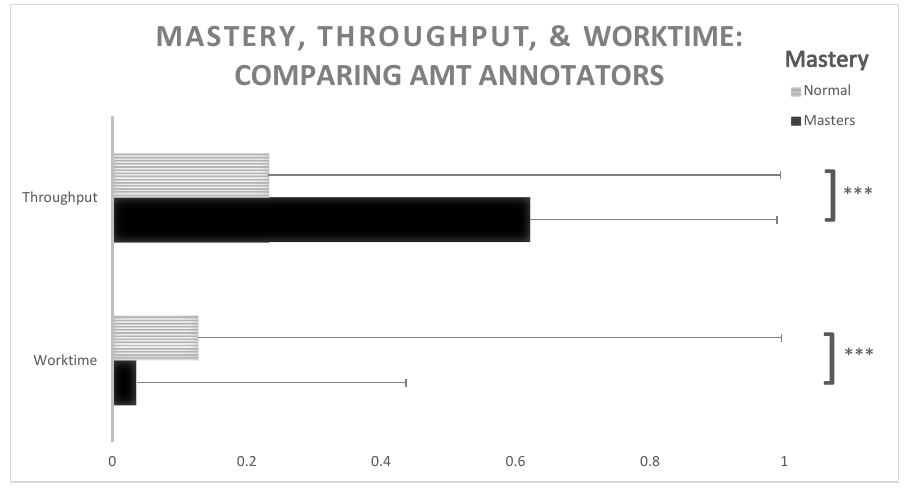}
 \includegraphics[width=.4\textwidth]{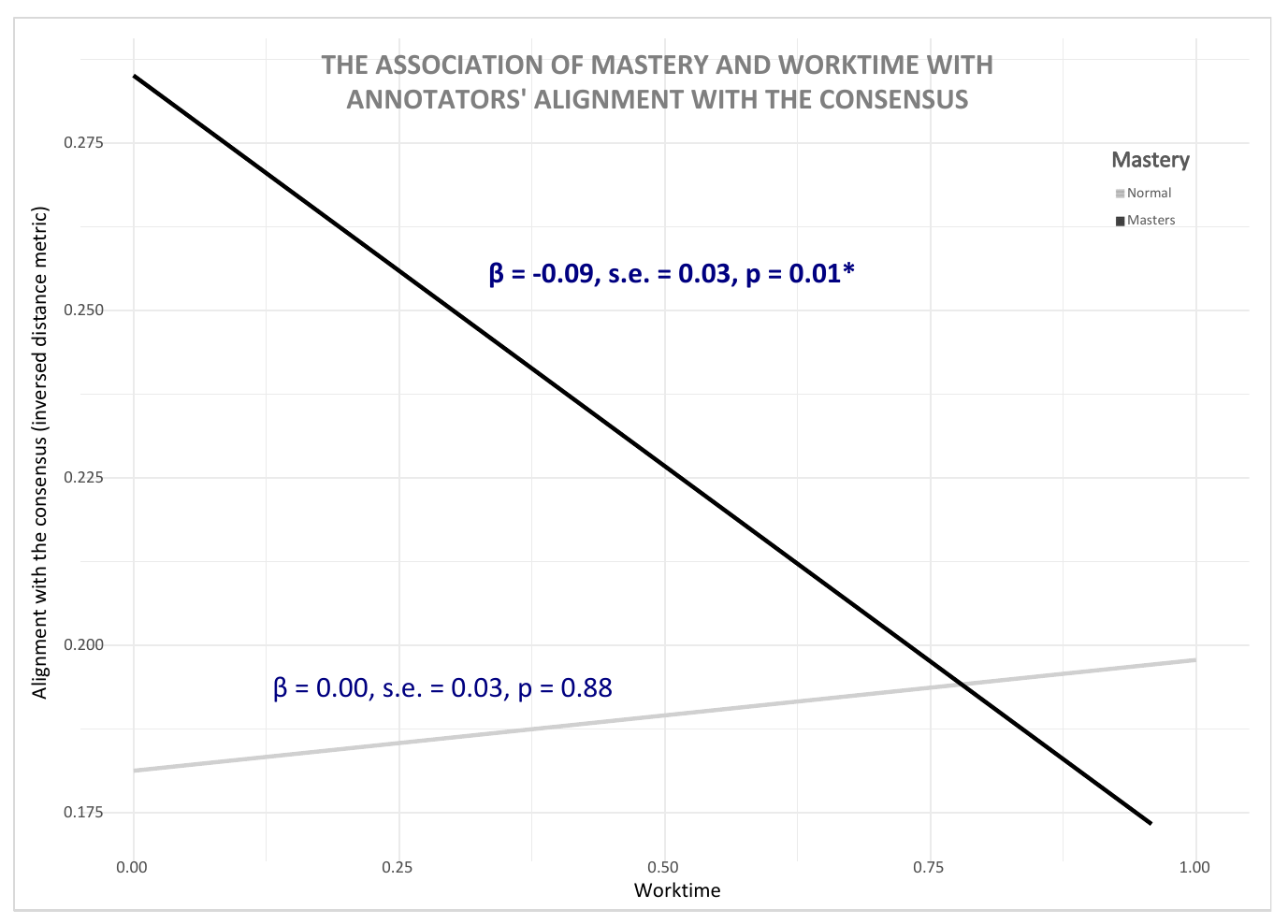}
 \caption{Comparative analysis of throughput and speed between master and normal Amazon Mechanical Turk (AMT) annotators. (a) Density plots of the distribution of worktime for annotators with and without Master's qualifications. (b) A comparison of the speed and throughput of annotators with and without Master's qualification (c) Master's annotators with higher worktime are less likely to align with the group than those with lower worktime.}
 \label{fig:mastervsnormal}
\end{figure}

\section{Discussion}
\label{sec:conclusion}
In this study, we introduce novel measures of annotator behavior beyond interpersonal agreement and demonstrate that incorporating annotator metadata, such as speed and throughput, and task metadata, such as text length, into the training corpus enhances predictive accuracy in user behavior prediction problems.

Drawing on Kahneman’s theory of attention~\cite{kahneman1973attention}, our findings underscore the importance of modeling annotator behavior as a function of cognitive resource allocation. Attention, a limited resource, is dynamically allocated based on task demands and effort, influencing annotation quality. Behavioral signals such as throughput and work time reflect attention depletion or disengagement, while text length captures task complexity and its cognitive toll. Our approach systematically addresses the challenges of annotation quality and enhances model performance across diverse datasets by leveraging signals of attention and effort.

Our results reveal that ensemble models enriched with these meta-features consistently outperform those without. The MSWEEM framework achieves significant gains, with a 14\% improvement over baseline ensemble methods on held-out data and a 12\% average improvement on a second dataset for a different task. These results highlight the generalizability of our approach, emphasizing the transferability of meta-feature utility across datasets and annotator cohorts. In the CLAff-Diplomacy dataset, features like work time and percentage agreement significantly improve performance, aligning with the cognitive demands of tasks requiring trust and deception evaluations. In contrast, in the CLAff-OffMyChest dataset, throughput and text length are more impactful due to the diversity and complexity of the text, where longer texts often induce fatigue and throughput reflects task familiarity. Annotator cohort differences further influence the impact of these features; for example, speed differences among Master's qualified workers require careful quality thresholds.

\subsection{Implications for Research}
Supervised machine learning models trained solely on textual properties often reflect biases in data collection, which can weaken performance~\cite{diaz2022crowdworksheets,lo2022your}. Annotation quality directly impacts model outcomes, and we demonstrate that incorporating behavioral meta-features leads to a 14\% improvement in classification accuracy for the target variable.

At an epistemological level, our findings advocate for the modeling of semantic properties alongside annotator metadata. External variables, such as task duration, prior priming, and social contexts, influence annotation quality and are reflected in features like throughput, work time, text length, and percentage agreement. Kahneman’s framework strengthens the justification for these features, emphasizing their role in addressing cognitive constraints.

Empirically, we validate the utility of behavioral signals in improving user behavior prediction models. Our approach innovatively uses existing metadata from platforms like Amazon Mechanical Turk, avoiding the need for additional monitoring software~\cite{goyal2018your}. While high throughput often correlates with annotator fatigue and poor quality, low work time signals potential speeding, although these associations vary across cohorts.

\subsection{Implications for Practice}
Facilitating crowdsourcing for data collection and developing relevant methodologies has been an important theme in the CSCW research community. Current lines of work focus on reducing biases and errors by comparing annotations~\cite{narimanzadeh2023crowdsourcing,kairam2016parting}.  Our work offers social impact through insights into how annotator behavior affects data quality.\footnote{The authors are sympathetic to the situation of crowdsourced workers. Many crowdsourced workers, especially MTurkers, consider AMT their primary source of income~\cite{ipeirotis2010demographics}. They are usually paid well below the minimum hourly wage~\cite{hara2018data}, which could be a possible reason for speeding through tasks.}

Our study on modeling behavioral patterns in annotations extends beyond NLP tasks and crowdsourced annotations. Specifically, our method offers a framework for handling data from online human subject studies, such as examining responses and engagement with texts. For instance, in research on alternative social media feeds, where users interact with posts, information and social overload may lead to speeding or fatigue that can result in inaccurate experimental outcomes~\cite{he2023cura}. To mitigate such risks, it may be necessary to calibrate results across participants by incorporating behavioral meta-features, enhancing the reliability of findings across studies.

With our insights, we propose a frame for processing online studies, particularly those that require human annotations. First, we need to decide when behavioral flaws such as speeding and fatigue become non-negligible such that the researcher needs to apply an ensemble framework that enriches the annotations with meta-features. If the annotation task is a behavioral outcome that thus requires cognitive understanding of a text (e.g., presence of deception, sentiment), and has some level of ambiguity (e.g., what reads as negative to one person may read as sarcasm to another), then it is a task suitable for MSWEEM. Conversely, if a task has deterministic answers, it may not need the MSWEEM architecture. Examples of such tasks are: ``Count the number of pronouns in the sentence.", or ``Identify the number of apples in the image".

After intuitively identifying the need of a MSWEEM architecture based on the type of task, the next step is to determine its need empirically. There are usually two annotation runs in an annotation run, a trial run where a subset of data is ran, and the full run where all the data is annotated~\cite{agirre2015semeval}. In a typical annotation setting, a single text is labeled by multiple annotators. During the trial run, the researcher should collect meta-features for each annotation and plot its distribution. If the distribution of the meta-features is a Normal distribution with a small variance, it means that the annotators are not really affected by behavioral signals when performing the task; it may be that the task does not require as much cognitive processing. If the distribution has a large variance or has a skewed distribution (e.g. Figure~\ref{fig:mastervsnormal}), then the MSWEEM architecture is required.

If the annotation task requires an MSWEEM setting for predictive labels, the following steps and guidelines can be followed: 

\begin{enumerate}
    \item Decide on the target variable: the target variable should be the variable that the research question desires to predict.
    \item Decide on the auxiliary variables: the auxiliary variables should be related semantically to the target variable. This step requires some knowledge of the dataset and references to linguistic or behavioral theory. Note that while we used four variables in both our datasets, the number of auxiliary variables need not only be limited to four. To aid in determining auxiliary variables $a_j$, some questions can be asked: What are related concepts to the target variable? What are related variables that can be derived from the dataset? What are the unique properties of the dataset, and can any of them be useful towards predicting the target variable? Can the variable $a_j$ be determined from the information given to the annotators, or does it need additional information (e.g. larger context, network information)? 
    \item Decide and experiment with probabilistic encoders: For each auxiliary variable, train and test individual classifiers where the inputs are the text and the outputs are the annotations (see Section~\ref{sec:indiv_classifiers}). Use the best performing classifier to build the probabilistic encoder layer of the ensemble. The probabilistic encoders need not be the same for each auxiliary variable.
\end{enumerate}

These steps thus set up the basic MSWEEM architecture for using an ensemble based model to perform prediction of behavioral outcomes.

\begin{table}[!ht]
\caption{\revision{Framework for Using the MSWEEM architecture}}
\label{tab:frameworkforpractice}
\begin{center}
\small
\begin{tabular}{p{0.5cm}p{4cm}p{7cm}}
\hline
\textbf{S/N} & \textbf{Step} & \textbf{Description} \\ \hline
1.1 & \footnotesize{\revision{Identify intuitively MSWEEM architecture is needed}} & \footnotesize{\revision{\textbf{Is the task a behavioral outcome that requires cognitive understanding?} If yes, go to 1.2. If no, MSWEEM not required \newline Examples: ``Annotate the presence of deception in the text"}} \\ 
1.2 & ~ & \footnotesize{\revision{\textbf{Is the task outcome  ambiguous?} Can two people annotate the same text in different ways? Are there more than one correct answer? \newline If yes, go to 1.3. If no, MSWEEM not required. }} \\
1.3 & ~ & \footnotesize{\revision{\textbf{Is the task deterministic?} Is there only one correct answer? \newline If yes, MSWEEM not required. If no, go to 1.4. \newline Examples: ``Count the number of pronouns in the sentence."}} \\ \hline 
1.4 & ~ & \footnotesize{\revision{\textbf{Is there a possible hierarchy of variables?} Are there multiple variables that can inform the task? Can these variables depend on each other?} \newline If yes, go to 2. If no, MSWEEM not required.} \\ 
\hline 
2 & \footnotesize{\revision{Decide on the target variable}} & \footnotesize{\revision{\textbf{What is the behavioral outcome to be predicted?}}} \newline Go to 3.1.\\ \hline 
3.1 & \footnotesize{\revision{Decide on the auxiliary variables}} & \footnotesize{\revision{\textbf{What are some related concepts to the target variable?} \newline Are there any linguistic or semantic concepts that are related to the target variable that aid in understanding the target variable?}} \newline Go to 3.2. \\
3.2 & ~ & \footnotesize{\revision{\textbf{What are the properties of the dataset that contribute to the target variable?} \newline What are unique properties of the dataset that are useful towards predicting the target variable? \newline Examples: presence of happiness for a r/happinesseveryday subreddit}} \newline Go to 3.3 \\ 
3.3 & ~ & \footnotesize{\revision{\textbf{Can the auxiliary variable be determined in the information given to the annotators?} If yes, go to 4. If no, it is not a good auxiliary variable. Go to 3.1 and find another auxiliary variable, or drop the auxiliary variable.}} \\ \hline 
4 & \footnotesize{\revision{Determine the probabilistic encoders for auxiliary variables}} & \footnotesize{\revision{For each auxiliary variable, train and test individual classifiers, where the inputs are the texts and the outputs are the auxiliary annotations. \newline Use the best performing classifier to build the probabilistic encoder layer of the ensemble. \newline Note: the probabilistic encoder need not be the same for each auxiliary variable. \newline Go to 5}} \\ \hline 
5 & \footnotesize{\revision{Run the MSWEEM Model}} & \\
\hline
\end{tabular}
\end{center}
\end{table}

Our frame is summarized in Table~\ref{tab:frameworkforpractice}. This frame illustrate how online annotations can be designed and administered to enhance annotation quality.

\subsection{Limitations and Future Work}
\label{sec:future}
While we see our data and model architecture as highly beneficial for the domain, there are some limitations to our work.
Our framework enhances ensemble model performance by leveraging annotator metadata from crowdsourcing platforms. While our experiments primarily use Amazon Mechanical Turk, the principles apply to other crowdsourcing platforms focused on human-labeled text classification. Not all platforms provide the same annotator performance signals. However, most platforms provide sufficient metadata to derive these from labels provided by the annotator (e.g., throughput, text length, percentage agreement) or estimate them (e.g., worktime), ensuring practical utility for this system. The ensemble approach to annotation requires auxiliary labels, yet the number of labels required is dependent on the ensemble setup. The annotator metadata is usually obtained during the annotation process itself. However, access to such data can vary, and factors like dataset size or label distribution may impact results. A promising direction could be to enrich ensemble frameworks with a small batch of known annotator-features when direct metadata is unavailable for the entire dataset is unavailable.

Workers' qualifications, and possibly self-reporting LLM use for annotations, are potentially useful quality meta-features to incorporate in future iterations of the ensemble framework. Future research could explore additional metadata available from crowdsourcing platforms and innovative ways to integrate this information. We also recommend that future scholars can perform further ablation analyses to discover the relationship between metadata representing annotator fatigue and speeding. This approach could extend to other areas, such as analyzing user engagement or content virality, by treating engagement signals as annotations. We encourage further investigation into complex weighting mechanisms to represent annotator behavior and comprehensive testing to understand how different annotator performance influences model performance. 

\section{Conclusion}
\label{sec:conclusion}
Inter-annotator agreement remains a challenge in creating high-quality training datasets for conversational analysis, impacting even advanced models like the Stanford Politeness model~\cite{danescu2013computational}, which suffer from low pairwise agreement. The limitations of traditional agreement measures like Fleiss and Cohen's Kappa, particularly for continuous data, necessitate innovative approaches~\cite{passonneau2014benefits}. 

The goal of this work was to gain an understanding for the annotator variability that behavioural datasets, which can contain subjective annotations, can contain. The Metadata-Sensitive Weighted-Encoding Ensemble Model (MSWEEM) bridges two critical gaps: enhancing knowledge representation through an ensemble approach for text input, significantly boosting specialized text classification performance, and addressing the issue of indeterminate label quality from crowdsourced annotations. Incorporating meta-features like annotator speed, text length, and throughput leads to superior model performance. This ensemble model is evaluated three fold: evaluation of model's performance gain, abalation analysis of the meta-features, dataset sizes, and over different annotator cohorts. The incorporation of meta-features provide another paradigm towards approach annotations: enabling researchers to adjust for quality variability, which therefore enhances model reliability in a time of uncertainty around the accuracy of crowdsourced labels. 

\section{Declaration}
\subsection{Availability of data and material}
The datasets are public and available at the resources linked in the main text. The code related to this study is at https://github.com/quarbby/diplomacy-betrayal.

\subsection{Competing interests}
The authors declare that they have no known competing financial interests or personal relationships that could have influenced the work reported in this paper.

\textbf{Acknowledgments:} This work was supported by a grant from the NUS Department of Communications and New Media and the Centre for Trusted Internet and Community. A special thanks to~\citet{peskov2020takes} and~\citet{niculae2015linguistic} for curating and sharing the original data. Wave 1 of annotations was also released as part of the CLAff-Diplomacy Shared Task~\cite{jaidkaeditorial}. We are also grateful to Akansha Gehlot, Ankita Patel, Fathima Vardha, Swastika Bhattacharya, and Sweta Kumari.

\bibliographystyle{ACM-Reference-Format}
\bibliography{references}


\begin{thebibliography}{74}


\ifx \showCODEN    \undefined \def \showCODEN     #1{\unskip}     \fi
\ifx \showDOI      \undefined \def \showDOI       #1{#1}\fi
\ifx \showISBNx    \undefined \def \showISBNx     #1{\unskip}     \fi
\ifx \showISBNxiii \undefined \def \showISBNxiii  #1{\unskip}     \fi
\ifx \showISSN     \undefined \def \showISSN      #1{\unskip}     \fi
\ifx \showLCCN     \undefined \def \showLCCN      #1{\unskip}     \fi
\ifx \shownote     \undefined \def \shownote      #1{#1}          \fi
\ifx \showarticletitle \undefined \def \showarticletitle #1{#1}   \fi
\ifx \showURL      \undefined \def \showURL       {\relax}        \fi
\providecommand\bibfield[2]{#2}
\providecommand\bibinfo[2]{#2}
\providecommand\natexlab[1]{#1}
\providecommand\showeprint[2][]{arXiv:#2}

\bibitem[Agirre et~al\mbox{.}(2015)]%
        {agirre2015semeval}
\bibfield{author}{\bibinfo{person}{Eneko Agirre}, \bibinfo{person}{Carmen Banea}, \bibinfo{person}{Claire Cardie}, \bibinfo{person}{Daniel Cer}, \bibinfo{person}{Mona Diab}, \bibinfo{person}{Aitor Gonzalez-Agirre}, \bibinfo{person}{Weiwei Guo}, \bibinfo{person}{Inigo Lopez-Gazpio}, \bibinfo{person}{Montse Maritxalar}, \bibinfo{person}{Rada Mihalcea}, {et~al\mbox{.}}} \bibinfo{year}{2015}\natexlab{}.
\newblock \showarticletitle{Semeval-2015 task 2: Semantic textual similarity, english, spanish and pilot on interpretability}. In \bibinfo{booktitle}{\emph{Proceedings of the 9th international workshop on semantic evaluation (SemEval 2015)}}. \bibinfo{pages}{252--263}.
\newblock


\bibitem[Artstein and Poesio(2008)]%
        {artstein2008inter}
\bibfield{author}{\bibinfo{person}{Ron Artstein} {and} \bibinfo{person}{Massimo Poesio}.} \bibinfo{year}{2008}\natexlab{}.
\newblock \showarticletitle{Inter-coder agreement for computational linguistics}.
\newblock \bibinfo{journal}{\emph{Computational linguistics}} \bibinfo{volume}{34}, \bibinfo{number}{4} (\bibinfo{year}{2008}), \bibinfo{pages}{555--596}.
\newblock


\bibitem[Bayerl and Paul(2011)]%
        {bayerl2011determines}
\bibfield{author}{\bibinfo{person}{Petra~Saskia Bayerl} {and} \bibinfo{person}{Karsten~Ingmar Paul}.} \bibinfo{year}{2011}\natexlab{}.
\newblock \showarticletitle{What determines inter-coder agreement in manual annotations? a meta-analytic investigation}.
\newblock \bibinfo{journal}{\emph{Computational Linguistics}} \bibinfo{volume}{37}, \bibinfo{number}{4} (\bibinfo{year}{2011}), \bibinfo{pages}{699--725}.
\newblock


\bibitem[Brownlee(2018)]%
        {brownlee2018better}
\bibfield{author}{\bibinfo{person}{Jason Brownlee}.} \bibinfo{year}{2018}\natexlab{}.
\newblock \bibinfo{booktitle}{\emph{Better deep learning: train faster, reduce overfitting, and make better predictions}}.
\newblock \bibinfo{publisher}{Machine Learning Mastery}.
\newblock


\bibitem[Carlson et~al\mbox{.}(2003)]%
        {carlson2003building}
\bibfield{author}{\bibinfo{person}{Lynn Carlson}, \bibinfo{person}{Daniel Marcu}, {and} \bibinfo{person}{Mary~Ellen Okurowski}.} \bibinfo{year}{2003}\natexlab{}.
\newblock \showarticletitle{Building a discourse-tagged corpus in the framework of rhetorical structure theory}.
\newblock \bibinfo{journal}{\emph{Current and new directions in discourse and dialogue}} (\bibinfo{year}{2003}), \bibinfo{pages}{85--112}.
\newblock


\bibitem[Chandler et~al\mbox{.}(2014)]%
        {chandler2014nonnaivete}
\bibfield{author}{\bibinfo{person}{Jesse Chandler}, \bibinfo{person}{Pam Mueller}, {and} \bibinfo{person}{Gabriele Paolacci}.} \bibinfo{year}{2014}\natexlab{}.
\newblock \showarticletitle{Nonna{\"\i}vet{\'e} among Amazon Mechanical Turk workers: Consequences and solutions for behavioral researchers}.
\newblock \bibinfo{journal}{\emph{Behavior research methods}}  \bibinfo{volume}{46} (\bibinfo{year}{2014}), \bibinfo{pages}{112--130}.
\newblock


\bibitem[Chatterjee and Bhattacharyya(2017)]%
        {CHATTERJEE2017138}
\bibfield{author}{\bibinfo{person}{Sujoy Chatterjee} {and} \bibinfo{person}{Malay Bhattacharyya}.} \bibinfo{year}{2017}\natexlab{}.
\newblock \showarticletitle{Judgment analysis of crowdsourced opinions using biclustering}.
\newblock \bibinfo{journal}{\emph{Information Sciences}}  \bibinfo{volume}{375} (\bibinfo{year}{2017}), \bibinfo{pages}{138--154}.
\newblock
\showISSN{0020-0255}
\urldef\tempurl%
\url{https://doi.org/10.1016/j.ins.2016.09.036}
\showDOI{\tempurl}


\bibitem[Chen et~al\mbox{.}(2022)]%
        {chen2022construction}
\bibfield{author}{\bibinfo{person}{Haihua Chen}, \bibinfo{person}{Lavinia~F Pieptea}, {and} \bibinfo{person}{Junhua Ding}.} \bibinfo{year}{2022}\natexlab{}.
\newblock \showarticletitle{Construction and evaluation of a high-quality corpus for legal intelligence using semiautomated approaches}.
\newblock \bibinfo{journal}{\emph{IEEE Transactions on Reliability}} \bibinfo{volume}{71}, \bibinfo{number}{2} (\bibinfo{year}{2022}), \bibinfo{pages}{657--673}.
\newblock


\bibitem[Chen and Yang(2021)]%
        {chen2021weakly}
\bibfield{author}{\bibinfo{person}{Jiaao Chen} {and} \bibinfo{person}{Diyi Yang}.} \bibinfo{year}{2021}\natexlab{}.
\newblock \showarticletitle{Weakly-Supervised Hierarchical Models for Predicting Persuasive Strategies in Good-faith Textual Requests}.
\newblock \bibinfo{journal}{\emph{arXiv preprint arXiv:2101.06351}} (\bibinfo{year}{2021}).
\newblock


\bibitem[Danescu-Niculescu-Mizil et~al\mbox{.}(2013)]%
        {danescu2013computational}
\bibfield{author}{\bibinfo{person}{Cristian Danescu-Niculescu-Mizil}, \bibinfo{person}{Moritz Sudhof}, \bibinfo{person}{Dan Jurafsky}, \bibinfo{person}{Jure Leskovec}, {and} \bibinfo{person}{Christopher Potts}.} \bibinfo{year}{2013}\natexlab{}.
\newblock \showarticletitle{A computational approach to politeness with application to social factors}.
\newblock \bibinfo{journal}{\emph{arXiv preprint arXiv:1306.6078}} (\bibinfo{year}{2013}).
\newblock


\bibitem[Devlin et~al\mbox{.}(2018)]%
        {devlin2018bert}
\bibfield{author}{\bibinfo{person}{Jacob Devlin}, \bibinfo{person}{Ming-Wei Chang}, \bibinfo{person}{Kenton Lee}, {and} \bibinfo{person}{Kristina Toutanova}.} \bibinfo{year}{2018}\natexlab{}.
\newblock \showarticletitle{BERT: Pre-training of Deep Bidirectional Transformers for Language Understanding}.
\newblock \bibinfo{journal}{\emph{arXiv preprint arXiv:1810.04805}} (\bibinfo{year}{2018}).
\newblock


\bibitem[D{\'\i}az et~al\mbox{.}(2022)]%
        {diaz2022crowdworksheets}
\bibfield{author}{\bibinfo{person}{Mark D{\'\i}az}, \bibinfo{person}{Ian Kivlichan}, \bibinfo{person}{Rachel Rosen}, \bibinfo{person}{Dylan Baker}, \bibinfo{person}{Razvan Amironesei}, \bibinfo{person}{Vinodkumar Prabhakaran}, {and} \bibinfo{person}{Emily Denton}.} \bibinfo{year}{2022}\natexlab{}.
\newblock \showarticletitle{Crowdworksheets: Accounting for individual and collective identities underlying crowdsourced dataset annotation}. In \bibinfo{booktitle}{\emph{Proceedings of the 2022 ACM Conference on Fairness, Accountability, and Transparency}}. \bibinfo{pages}{2342--2351}.
\newblock


\bibitem[Difallah et~al\mbox{.}(2018)]%
        {difallah2018demographics}
\bibfield{author}{\bibinfo{person}{Djellel Difallah}, \bibinfo{person}{Elena Filatova}, {and} \bibinfo{person}{Panos Ipeirotis}.} \bibinfo{year}{2018}\natexlab{}.
\newblock \showarticletitle{Demographics and dynamics of mechanical turk workers}. In \bibinfo{booktitle}{\emph{Proceedings of the eleventh ACM international conference on web search and data mining}}. \bibinfo{pages}{135--143}.
\newblock


\bibitem[Douglas et~al\mbox{.}(2023)]%
        {douglas2023data}
\bibfield{author}{\bibinfo{person}{Benjamin~D Douglas}, \bibinfo{person}{Patrick~J Ewell}, {and} \bibinfo{person}{Markus Brauer}.} \bibinfo{year}{2023}\natexlab{}.
\newblock \showarticletitle{Data quality in online human-subjects research: Comparisons between MTurk, Prolific, CloudResearch, Qualtrics, and SONA}.
\newblock \bibinfo{journal}{\emph{Plos one}} \bibinfo{volume}{18}, \bibinfo{number}{3} (\bibinfo{year}{2023}), \bibinfo{pages}{e0279720}.
\newblock


\bibitem[Eickhoff and de~Vries(2013)]%
        {eickhoff2013increasing}
\bibfield{author}{\bibinfo{person}{Carsten Eickhoff} {and} \bibinfo{person}{Arjen~P de Vries}.} \bibinfo{year}{2013}\natexlab{}.
\newblock \showarticletitle{Increasing cheat robustness of crowdsourcing tasks}.
\newblock \bibinfo{journal}{\emph{Information retrieval}}  \bibinfo{volume}{16} (\bibinfo{year}{2013}), \bibinfo{pages}{121--137}.
\newblock


\bibitem[Fornaciari et~al\mbox{.}(2021)]%
        {fornaciari2021beyond}
\bibfield{author}{\bibinfo{person}{Tommaso Fornaciari}, \bibinfo{person}{Alexandra Uma}, \bibinfo{person}{Silviu Paun}, \bibinfo{person}{Barbara Plank}, \bibinfo{person}{Dirk Hovy}, {and} \bibinfo{person}{Massimo Poesio}.} \bibinfo{year}{2021}\natexlab{}.
\newblock \showarticletitle{Beyond Black \& White: Leveraging Annotator Disagreement via Soft-Label Multi-Task Learning}. In \bibinfo{booktitle}{\emph{Proceedings of the 2021 Conference of the North American Chapter of the Association for Computational Linguistics: Human Language Technologies}}. \bibinfo{pages}{2591--2597}.
\newblock


\bibitem[Ganaie et~al\mbox{.}(2022)]%
        {ganaie2022ensemble}
\bibfield{author}{\bibinfo{person}{Mudasir~A Ganaie}, \bibinfo{person}{Minghui Hu}, \bibinfo{person}{AK Malik}, \bibinfo{person}{M Tanveer}, {and} \bibinfo{person}{PN Suganthan}.} \bibinfo{year}{2022}\natexlab{}.
\newblock \showarticletitle{Ensemble deep learning: A review}.
\newblock \bibinfo{journal}{\emph{Engineering Applications of Artificial Intelligence}}  \bibinfo{volume}{115} (\bibinfo{year}{2022}), \bibinfo{pages}{105151}.
\newblock


\bibitem[Geva et~al\mbox{.}(2019)]%
        {geva2019we}
\bibfield{author}{\bibinfo{person}{Mor Geva}, \bibinfo{person}{Yoav Goldberg}, {and} \bibinfo{person}{Jonathan Berant}.} \bibinfo{year}{2019}\natexlab{}.
\newblock \showarticletitle{Are we modeling the task or the annotator? an investigation of annotator bias in natural language understanding datasets}.
\newblock \bibinfo{journal}{\emph{arXiv preprint arXiv:1908.07898}} (\bibinfo{year}{2019}).
\newblock


\bibitem[Goyal et~al\mbox{.}(2018)]%
        {goyal2018your}
\bibfield{author}{\bibinfo{person}{Tanya Goyal}, \bibinfo{person}{Tyler McDonnell}, \bibinfo{person}{Mucahid Kutlu}, \bibinfo{person}{Tamer Elsayed}, {and} \bibinfo{person}{Matthew Lease}.} \bibinfo{year}{2018}\natexlab{}.
\newblock \showarticletitle{Your behavior signals your reliability: Modeling crowd behavioral traces to ensure quality relevance annotations}. In \bibinfo{booktitle}{\emph{Proceedings of the AAAI Conference on Human Computation and Crowdsourcing}}, Vol.~\bibinfo{volume}{6}. \bibinfo{pages}{41--49}.
\newblock


\bibitem[Guo et~al\mbox{.}(2019)]%
        {10.48550/arxiv.1909.12911}
\bibfield{author}{\bibinfo{person}{X. Guo}, \bibinfo{person}{L. Polania}, \bibinfo{person}{B. Zhu}, \bibinfo{person}{C. Boncelet}, {and} \bibinfo{person}{K. Barner}.} \bibinfo{year}{2019}\natexlab{}.
\newblock \showarticletitle{Graph neural networks for image understanding based on multiple cues: group emotion recognition and event recognition as use cases}.
\newblock  (\bibinfo{year}{2019}).
\newblock
\urldef\tempurl%
\url{https://doi.org/10.48550/arxiv.1909.12911}
\showDOI{\tempurl}


\bibitem[Hara et~al\mbox{.}(2018)]%
        {hara2018data}
\bibfield{author}{\bibinfo{person}{Kotaro Hara}, \bibinfo{person}{Abigail Adams}, \bibinfo{person}{Kristy Milland}, \bibinfo{person}{Saiph Savage}, \bibinfo{person}{Chris Callison-Burch}, {and} \bibinfo{person}{Jeffrey~P Bigham}.} \bibinfo{year}{2018}\natexlab{}.
\newblock \showarticletitle{A data-driven analysis of workers' earnings on Amazon Mechanical Turk}. In \bibinfo{booktitle}{\emph{Proceedings of the 2018 CHI conference on human factors in computing systems}}. \bibinfo{pages}{1--14}.
\newblock


\bibitem[He et~al\mbox{.}(2023)]%
        {he2023cura}
\bibfield{author}{\bibinfo{person}{Wanrong He}, \bibinfo{person}{Mitchell~L Gordon}, \bibinfo{person}{Lindsay Popowski}, {and} \bibinfo{person}{Michael~S Bernstein}.} \bibinfo{year}{2023}\natexlab{}.
\newblock \showarticletitle{Cura: Curation at Social Media Scale}.
\newblock \bibinfo{journal}{\emph{Proceedings of the ACM on Human-Computer Interaction}} \bibinfo{volume}{7}, \bibinfo{number}{CSCW2} (\bibinfo{year}{2023}), \bibinfo{pages}{1--33}.
\newblock


\bibitem[He et~al\mbox{.}(2024)]%
        {he2024if}
\bibfield{author}{\bibinfo{person}{Zeyu He}, \bibinfo{person}{Chieh-Yang Huang}, \bibinfo{person}{Chien-Kuang~Cornelia Ding}, \bibinfo{person}{Shaurya Rohatgi}, {and} \bibinfo{person}{Ting-Hao'Kenneth' Huang}.} \bibinfo{year}{2024}\natexlab{}.
\newblock \showarticletitle{If in a Crowdsourced Data Annotation Pipeline, a GPT-4}.
\newblock \bibinfo{journal}{\emph{arXiv preprint arXiv:2402.16795}} (\bibinfo{year}{2024}).
\newblock


\bibitem[Hochreiter and Schmidhuber(1997)]%
        {hochreiter1997long}
\bibfield{author}{\bibinfo{person}{Sepp Hochreiter} {and} \bibinfo{person}{J{\"u}rgen Schmidhuber}.} \bibinfo{year}{1997}\natexlab{}.
\newblock \showarticletitle{Long short-term memory}.
\newblock \bibinfo{journal}{\emph{Neural computation}} \bibinfo{volume}{9}, \bibinfo{number}{8} (\bibinfo{year}{1997}), \bibinfo{pages}{1735--1780}.
\newblock


\bibitem[Hovy et~al\mbox{.}(2013)]%
        {hovy-etal-2013-learning}
\bibfield{author}{\bibinfo{person}{Dirk Hovy}, \bibinfo{person}{Taylor Berg-Kirkpatrick}, \bibinfo{person}{Ashish Vaswani}, {and} \bibinfo{person}{Eduard Hovy}.} \bibinfo{year}{2013}\natexlab{}.
\newblock \showarticletitle{Learning Whom to Trust with {MACE}}. In \bibinfo{booktitle}{\emph{Proceedings of the 2013 Conference of the North {A}merican Chapter of the Association for Computational Linguistics: Human Language Technologies}}, \bibfield{editor}{\bibinfo{person}{Lucy Vanderwende}, \bibinfo{person}{Hal Daum{\'e}~III}, {and} \bibinfo{person}{Katrin Kirchhoff}} (Eds.). \bibinfo{publisher}{Association for Computational Linguistics}, \bibinfo{address}{Atlanta, Georgia}, \bibinfo{pages}{1120--1130}.
\newblock
\urldef\tempurl%
\url{https://aclanthology.org/N13-1132}
\showURL{%
\tempurl}


\bibitem[Hovy and Yang(2021)]%
        {hovy2021importance}
\bibfield{author}{\bibinfo{person}{Dirk Hovy} {and} \bibinfo{person}{Diyi Yang}.} \bibinfo{year}{2021}\natexlab{}.
\newblock \showarticletitle{The importance of modeling social factors of language: Theory and practice}. In \bibinfo{booktitle}{\emph{Proceedings of the 2021 Conference of the North American Chapter of the Association for Computational Linguistics: Human Language Technologies}}. \bibinfo{pages}{588--602}.
\newblock


\bibitem[Hovy and Lavid(2010)]%
        {hovy2010towards}
\bibfield{author}{\bibinfo{person}{Eduard Hovy} {and} \bibinfo{person}{Julia Lavid}.} \bibinfo{year}{2010}\natexlab{}.
\newblock \showarticletitle{Towards a ‘Science’of Corpus Annotation: A New Methodological Challenge for Corpus Linguistics}.
\newblock \bibinfo{journal}{\emph{INTERNATIONAL JOURNAL OF TRANSLATION}} \bibinfo{volume}{22}, \bibinfo{number}{1} (\bibinfo{year}{2010}).
\newblock


\bibitem[Hsueh et~al\mbox{.}(2009a)]%
        {hsueh2009data}
\bibfield{author}{\bibinfo{person}{Pei-Yun Hsueh}, \bibinfo{person}{Prem Melville}, {and} \bibinfo{person}{Vikas Sindhwani}.} \bibinfo{year}{2009}\natexlab{a}.
\newblock \showarticletitle{Data quality from crowdsourcing: a study of annotation selection criteria}. In \bibinfo{booktitle}{\emph{Proceedings of the NAACL HLT 2009 workshop on active learning for natural language processing}}. \bibinfo{pages}{27--35}.
\newblock


\bibitem[Hsueh et~al\mbox{.}(2009b)]%
        {hsueh-etal-2009-data}
\bibfield{author}{\bibinfo{person}{Pei-Yun Hsueh}, \bibinfo{person}{Prem Melville}, {and} \bibinfo{person}{Vikas Sindhwani}.} \bibinfo{year}{2009}\natexlab{b}.
\newblock \showarticletitle{Data Quality from Crowdsourcing: A Study of Annotation Selection Criteria}. In \bibinfo{booktitle}{\emph{Proceedings of the {NAACL} {HLT} 2009 Workshop on Active Learning for Natural Language Processing}}. \bibinfo{publisher}{Association for Computational Linguistics}, \bibinfo{address}{Boulder, Colorado}, \bibinfo{pages}{27--35}.
\newblock
\urldef\tempurl%
\url{https://www.aclweb.org/anthology/W09-1904}
\showURL{%
\tempurl}


\bibitem[Ipeirotis(2010)]%
        {ipeirotis2010demographics}
\bibfield{author}{\bibinfo{person}{Panagiotis~G Ipeirotis}.} \bibinfo{year}{2010}\natexlab{}.
\newblock \showarticletitle{Demographics of mechanical turk}.
\newblock  (\bibinfo{year}{2010}).
\newblock


\bibitem[Ipeirotis et~al\mbox{.}(2010)]%
        {ipeirotis2010quality}
\bibfield{author}{\bibinfo{person}{Panagiotis~G Ipeirotis}, \bibinfo{person}{Foster Provost}, {and} \bibinfo{person}{Jing Wang}.} \bibinfo{year}{2010}\natexlab{}.
\newblock \showarticletitle{Quality management on amazon mechanical turk}. In \bibinfo{booktitle}{\emph{Proceedings of the ACM SIGKDD workshop on human computation}}. \bibinfo{pages}{64--67}.
\newblock


\bibitem[Jaidka et~al\mbox{.}(2024)]%
        {jaidka2024takes}
\bibfield{author}{\bibinfo{person}{Kokil Jaidka}, \bibinfo{person}{Hansin Ahuja}, {and} \bibinfo{person}{Lynnette Hui~Xian Ng}.} \bibinfo{year}{2024}\natexlab{}.
\newblock \showarticletitle{It Takes Two to Negotiate: Modeling Social Exchange in Online Multiplayer Games}.
\newblock \bibinfo{journal}{\emph{Proceedings of the ACM on Human-Computer Interaction}} \bibinfo{volume}{8}, \bibinfo{number}{CSCW1} (\bibinfo{year}{2024}), \bibinfo{pages}{1--22}.
\newblock


\bibitem[Jaidka et~al\mbox{.}(2021a)]%
        {jaidka2021wikitalkedit}
\bibfield{author}{\bibinfo{person}{Kokil Jaidka}, \bibinfo{person}{Andrea Ceolin}, \bibinfo{person}{Iknoor Singh}, \bibinfo{person}{Niyati Chhaya}, {and} \bibinfo{person}{Lyle Ungar}.} \bibinfo{year}{2021}\natexlab{a}.
\newblock \showarticletitle{WikiTalkEdit: A Dataset for modeling Editors’ behaviors on Wikipedia}. In \bibinfo{booktitle}{\emph{Proceedings of the 2021 Conference of the North American Chapter of the Association for Computational Linguistics: Human Language Technologies}}. \bibinfo{pages}{2191--2200}.
\newblock


\bibitem[Jaidka et~al\mbox{.}(2021b)]%
        {jaidkaeditorial}
\bibfield{author}{\bibinfo{person}{Kokil Jaidka}, \bibinfo{person}{Niyati Chhaya}, \bibinfo{person}{Lyle Ungar}, \bibinfo{person}{Jennifer Healey}, {and} \bibinfo{person}{Atanu Sinha}.} \bibinfo{year}{2021}\natexlab{b}.
\newblock \showarticletitle{Editorial for the 4th AAAI-21 Workshop on Affective Content Analysis}.
\newblock  (\bibinfo{year}{2021}).
\newblock


\bibitem[Jaidka et~al\mbox{.}(2020a)]%
        {jaidka2020report}
\bibfield{author}{\bibinfo{person}{Kokil Jaidka}, \bibinfo{person}{Iknoor Singh}, \bibinfo{person}{Jiahui Liu}, \bibinfo{person}{Niyati Chhaya}, {and} \bibinfo{person}{Lyle Ungar}.} \bibinfo{year}{2020}\natexlab{a}.
\newblock \showarticletitle{A report of the CL-Aff OffMyChest Shared Task: Modeling Supportiveness and Disclosure.}. In \bibinfo{booktitle}{\emph{AffCon@ AAAI}}. \bibinfo{pages}{118--129}.
\newblock


\bibitem[Jaidka et~al\mbox{.}(2020b)]%
        {DBLP:conf/aaai/JaidkaSLCU20}
\bibfield{author}{\bibinfo{person}{Kokil Jaidka}, \bibinfo{person}{Iknoor Singh}, \bibinfo{person}{Jiahui Liu}, \bibinfo{person}{Niyati Chhaya}, {and} \bibinfo{person}{Lyle Ungar}.} \bibinfo{year}{2020}\natexlab{b}.
\newblock \showarticletitle{A report of the CL-Aff OffMyChest Shared Task: Modeling Supportiveness and Disclosure}. In \bibinfo{booktitle}{\emph{Proceedings of the 3rd Workshop on Affective Content Analysis (AffCon 2020) co-located with Thirty-Fourth {AAAI} Conference on Artificial Intelligence {(AAAI} 2020), New York, USA, February 7, 2020}} \emph{(\bibinfo{series}{{CEUR} Workshop Proceedings}, Vol.~\bibinfo{volume}{2614})}, \bibfield{editor}{\bibinfo{person}{Niyati Chhaya}, \bibinfo{person}{Kokil Jaidka}, \bibinfo{person}{Jennifer Healey}, \bibinfo{person}{Lyle Ungar}, {and} \bibinfo{person}{Atanu~R. Sinha}} (Eds.). \bibinfo{publisher}{CEUR-WS.org}, \bibinfo{pages}{118--129}.
\newblock
\urldef\tempurl%
\url{http://ceur-ws.org/Vol-2614/AffCon20\_session3\_claffoverview.pdf}
\showURL{%
\tempurl}


\bibitem[Kahneman(1973)]%
        {kahneman1973attention}
\bibfield{author}{\bibinfo{person}{Daniel Kahneman}.} \bibinfo{year}{1973}\natexlab{}.
\newblock \bibinfo{booktitle}{\emph{Attention and Effort}}.
\newblock \bibinfo{publisher}{Prentice-Hall}.
\newblock


\bibitem[Kairam and Heer(2016)]%
        {kairam2016parting}
\bibfield{author}{\bibinfo{person}{Sanjay Kairam} {and} \bibinfo{person}{Jeffrey Heer}.} \bibinfo{year}{2016}\natexlab{}.
\newblock \showarticletitle{Parting crowds: Characterizing divergent interpretations in crowdsourced annotation tasks}. In \bibinfo{booktitle}{\emph{Proceedings of the 19th ACM Conference on Computer-Supported Cooperative Work \& Social Computing}}. \bibinfo{pages}{1637--1648}.
\newblock


\bibitem[Kazai et~al\mbox{.}(2011)]%
        {kazai2011worker}
\bibfield{author}{\bibinfo{person}{Gabriella Kazai}, \bibinfo{person}{Jaap Kamps}, {and} \bibinfo{person}{Natasa Milic-Frayling}.} \bibinfo{year}{2011}\natexlab{}.
\newblock \showarticletitle{Worker types and personality traits in crowdsourcing relevance labels}. In \bibinfo{booktitle}{\emph{Proceedings of the 20th ACM international conference on Information and knowledge management}}. \bibinfo{pages}{1941--1944}.
\newblock


\bibitem[Kim(2014)]%
        {kim-2014-convolutional}
\bibfield{author}{\bibinfo{person}{Yoon Kim}.} \bibinfo{year}{2014}\natexlab{}.
\newblock \showarticletitle{Convolutional Neural Networks for Sentence Classification}. In \bibinfo{booktitle}{\emph{Proceedings of the 2014 Conference on Empirical Methods in Natural Language Processing ({EMNLP})}}. \bibinfo{publisher}{Association for Computational Linguistics}, \bibinfo{address}{Doha, Qatar}, \bibinfo{pages}{1746--1751}.
\newblock
\urldef\tempurl%
\url{https://doi.org/10.3115/v1/D14-1181}
\showDOI{\tempurl}


\bibitem[Litman et~al\mbox{.}(2015)]%
        {litman2015relationship}
\bibfield{author}{\bibinfo{person}{Leib Litman}, \bibinfo{person}{Jonathan Robinson}, {and} \bibinfo{person}{Cheskie Rosenzweig}.} \bibinfo{year}{2015}\natexlab{}.
\newblock \showarticletitle{The relationship between motivation, monetary compensation, and data quality among US-and India-based workers on Mechanical Turk}.
\newblock \bibinfo{journal}{\emph{Behavior research methods}} \bibinfo{volume}{47}, \bibinfo{number}{2} (\bibinfo{year}{2015}), \bibinfo{pages}{519--528}.
\newblock


\bibitem[Liu et~al\mbox{.}(2015)]%
        {liu-etal-2015-fine}
\bibfield{author}{\bibinfo{person}{Pengfei Liu}, \bibinfo{person}{Shafiq Joty}, {and} \bibinfo{person}{Helen Meng}.} \bibinfo{year}{2015}\natexlab{}.
\newblock \showarticletitle{Fine-grained Opinion Mining with Recurrent Neural Networks and Word Embeddings}. In \bibinfo{booktitle}{\emph{Proceedings of the 2015 Conference on Empirical Methods in Natural Language Processing}}. \bibinfo{publisher}{Association for Computational Linguistics}, \bibinfo{address}{Lisbon, Portugal}, \bibinfo{pages}{1433--1443}.
\newblock
\urldef\tempurl%
\url{https://doi.org/10.18653/v1/D15-1168}
\showDOI{\tempurl}


\bibitem[Liu et~al\mbox{.}(2019)]%
        {DBLP:journals/corr/abs-1907-11692}
\bibfield{author}{\bibinfo{person}{Yinhan Liu}, \bibinfo{person}{Myle Ott}, \bibinfo{person}{Naman Goyal}, \bibinfo{person}{Jingfei Du}, \bibinfo{person}{Mandar Joshi}, \bibinfo{person}{Danqi Chen}, \bibinfo{person}{Omer Levy}, \bibinfo{person}{Mike Lewis}, \bibinfo{person}{Luke Zettlemoyer}, {and} \bibinfo{person}{Veselin Stoyanov}.} \bibinfo{year}{2019}\natexlab{}.
\newblock \showarticletitle{RoBERTa: {A} Robustly Optimized {BERT} Pretraining Approach}.
\newblock \bibinfo{journal}{\emph{CoRR}}  \bibinfo{volume}{abs/1907.11692} (\bibinfo{year}{2019}).
\newblock
\showeprint[arxiv]{1907.11692}
\urldef\tempurl%
\url{http://arxiv.org/abs/1907.11692}
\showURL{%
\tempurl}


\bibitem[Lo and Lim(2022)]%
        {lo2022your}
\bibfield{author}{\bibinfo{person}{Pei-Chi Lo} {and} \bibinfo{person}{Ee-Peng Lim}.} \bibinfo{year}{2022}\natexlab{}.
\newblock \showarticletitle{Your Cursor Reveals: On Analyzing Workers’ Browsing Behavior and Annotation Quality In Crowdsourcing Tasks}.
\newblock \bibinfo{journal}{\emph{IEEE Access}} (\bibinfo{year}{2022}).
\newblock


\bibitem[Lourentzou(2019)]%
        {lourentzou2019data}
\bibfield{author}{\bibinfo{person}{Ismini Lourentzou}.} \bibinfo{year}{2019}\natexlab{}.
\newblock \showarticletitle{Data quality in the deep learning ERA: Active semi-supervised learning and text normalization for natural language understanding}.
\newblock  (\bibinfo{year}{2019}).
\newblock


\bibitem[Narimanzadeh et~al\mbox{.}(2023)]%
        {narimanzadeh2023crowdsourcing}
\bibfield{author}{\bibinfo{person}{Hasti Narimanzadeh}, \bibinfo{person}{Arash Badie-Modiri}, \bibinfo{person}{Iuliia~G Smirnova}, {and} \bibinfo{person}{Ted Hsuan~Yun Chen}.} \bibinfo{year}{2023}\natexlab{}.
\newblock \showarticletitle{Crowdsourcing subjective annotations using pairwise comparisons reduces bias and error compared to the majority-vote method}.
\newblock \bibinfo{journal}{\emph{Proceedings of the ACM on Human-Computer Interaction}} \bibinfo{volume}{7}, \bibinfo{number}{CSCW2} (\bibinfo{year}{2023}), \bibinfo{pages}{1--29}.
\newblock


\bibitem[Ng and Carley(2023)]%
        {ng2023botbuster}
\bibfield{author}{\bibinfo{person}{Lynnette Hui~Xian Ng} {and} \bibinfo{person}{Kathleen~M Carley}.} \bibinfo{year}{2023}\natexlab{}.
\newblock \showarticletitle{Botbuster: Multi-platform bot detection using a mixture of experts}. In \bibinfo{booktitle}{\emph{Proceedings of the International AAAI Conference on Web and Social Media}}, Vol.~\bibinfo{volume}{17}. \bibinfo{pages}{686--697}.
\newblock


\bibitem[Niculae et~al\mbox{.}(2015)]%
        {niculae2015linguistic}
\bibfield{author}{\bibinfo{person}{Vlad Niculae}, \bibinfo{person}{Srijan Kumar}, \bibinfo{person}{Jordan Boyd-Graber}, {and} \bibinfo{person}{Cristian Danescu-Niculescu-Mizil}.} \bibinfo{year}{2015}\natexlab{}.
\newblock \showarticletitle{Linguistic Harbingers of Betrayal: A Case Study on an Online Strategy Game}. In \bibinfo{booktitle}{\emph{Proceedings of the 53rd Annual Meeting of the Association for Computational Linguistics and the 7th International Joint Conference on Natural Language Processing (Volume 1: Long Papers)}}. \bibinfo{pages}{1650--1659}.
\newblock


\bibitem[Parikh et~al\mbox{.}(2019)]%
        {parikh2019multi}
\bibfield{author}{\bibinfo{person}{Pulkit Parikh}, \bibinfo{person}{Harika Abburi}, \bibinfo{person}{Pinkesh Badjatiya}, \bibinfo{person}{Radhika Krishnan}, \bibinfo{person}{Niyati Chhaya}, \bibinfo{person}{Manish Gupta}, {and} \bibinfo{person}{Vasudeva Varma}.} \bibinfo{year}{2019}\natexlab{}.
\newblock \showarticletitle{Multi-label Categorization of Accounts of Sexism using a Neural Framework}. In \bibinfo{booktitle}{\emph{Proceedings of the 2019 Conference on Empirical Methods in Natural Language Processing and the 9th International Joint Conference on Natural Language Processing (EMNLP-IJCNLP)}}. \bibinfo{pages}{1642--1652}.
\newblock


\bibitem[Passonneau and Carpenter(2014)]%
        {passonneau2014benefits}
\bibfield{author}{\bibinfo{person}{Rebecca~J Passonneau} {and} \bibinfo{person}{Bob Carpenter}.} \bibinfo{year}{2014}\natexlab{}.
\newblock \showarticletitle{The benefits of a model of annotation}.
\newblock \bibinfo{journal}{\emph{Transactions of the Association for Computational Linguistics}}  \bibinfo{volume}{2} (\bibinfo{year}{2014}), \bibinfo{pages}{311--326}.
\newblock


\bibitem[Paun et~al\mbox{.}(2018)]%
        {paun2018comparing}
\bibfield{author}{\bibinfo{person}{Silviu Paun}, \bibinfo{person}{Bob Carpenter}, \bibinfo{person}{Jon Chamberlain}, \bibinfo{person}{Dirk Hovy}, \bibinfo{person}{Udo Kruschwitz}, {and} \bibinfo{person}{Massimo Poesio}.} \bibinfo{year}{2018}\natexlab{}.
\newblock \showarticletitle{Comparing bayesian models of annotation}.
\newblock \bibinfo{journal}{\emph{Transactions of the Association for Computational Linguistics}}  \bibinfo{volume}{6} (\bibinfo{year}{2018}), \bibinfo{pages}{571--585}.
\newblock


\bibitem[Perikos and Hatzilygeroudis(2016)]%
        {PERIKOS2016191}
\bibfield{author}{\bibinfo{person}{Isidoros Perikos} {and} \bibinfo{person}{Ioannis Hatzilygeroudis}.} \bibinfo{year}{2016}\natexlab{}.
\newblock \showarticletitle{Recognizing emotions in text using ensemble of classifiers}.
\newblock \bibinfo{journal}{\emph{Engineering Applications of Artificial Intelligence}}  \bibinfo{volume}{51} (\bibinfo{year}{2016}), \bibinfo{pages}{191--201}.
\newblock
\showISSN{0952-1976}
\urldef\tempurl%
\url{https://doi.org/10.1016/j.engappai.2016.01.012}
\showDOI{\tempurl}
\newblock
\shownote{Mining the Humanities: Technologies and Applications}.


\bibitem[Peskov et~al\mbox{.}(2020a)]%
        {peskov-etal-2020-takes}
\bibfield{author}{\bibinfo{person}{Denis Peskov}, \bibinfo{person}{Benny Cheng}, \bibinfo{person}{Ahmed Elgohary}, \bibinfo{person}{Joe Barrow}, \bibinfo{person}{Cristian Danescu-Niculescu-Mizil}, {and} \bibinfo{person}{Jordan Boyd-Graber}.} \bibinfo{year}{2020}\natexlab{a}.
\newblock \showarticletitle{It Takes Two to Lie: One to Lie, and One to Listen}. In \bibinfo{booktitle}{\emph{Proceedings of the 58th Annual Meeting of the Association for Computational Linguistics}}. \bibinfo{publisher}{Association for Computational Linguistics}, \bibinfo{address}{Online}, \bibinfo{pages}{3811--3854}.
\newblock
\urldef\tempurl%
\url{https://doi.org/10.18653/v1/2020.acl-main.353}
\showDOI{\tempurl}


\bibitem[Peskov et~al\mbox{.}(2020b)]%
        {peskov2020takes}
\bibfield{author}{\bibinfo{person}{Denis Peskov}, \bibinfo{person}{Benny Cheng}, \bibinfo{person}{Ahmed Elgohary}, \bibinfo{person}{Joe Barrow}, \bibinfo{person}{Cristian Danescu-Niculescu-Mizil}, {and} \bibinfo{person}{Jordan Boyd-Graber}.} \bibinfo{year}{2020}\natexlab{b}.
\newblock \showarticletitle{It Takes Two to Lie: One to Lie, and One to Listen}. In \bibinfo{booktitle}{\emph{Proceedings of the 58th Annual Meeting of the Association for Computational Linguistics}}. \bibinfo{pages}{3811--3854}.
\newblock


\bibitem[Rodrigues and Pereira(2018)]%
        {rodrigues2018deep}
\bibfield{author}{\bibinfo{person}{Filipe Rodrigues} {and} \bibinfo{person}{Francisco Pereira}.} \bibinfo{year}{2018}\natexlab{}.
\newblock \showarticletitle{Deep learning from crowds}. In \bibinfo{booktitle}{\emph{Proceedings of the AAAI Conference on Artificial Intelligence}}, Vol.~\bibinfo{volume}{32}.
\newblock


\bibitem[Saeed et~al\mbox{.}(2018)]%
        {saeed2018impact}
\bibfield{author}{\bibinfo{person}{Nagwa~MK Saeed}, \bibinfo{person}{Nivin~A Helal}, \bibinfo{person}{Nagwa~L Badr}, {and} \bibinfo{person}{Tarek~F Gharib}.} \bibinfo{year}{2018}\natexlab{}.
\newblock \showarticletitle{The impact of spam reviews on feature-based sentiment analysis}. In \bibinfo{booktitle}{\emph{2018 13th International Conference on Computer Engineering and Systems (ICCES)}}. IEEE, \bibinfo{pages}{633--639}.
\newblock


\bibitem[Saeed et~al\mbox{.}(2022)]%
        {saeed2022ensemble}
\bibfield{author}{\bibinfo{person}{Radwa~MK Saeed}, \bibinfo{person}{Sherine Rady}, {and} \bibinfo{person}{Tarek~F Gharib}.} \bibinfo{year}{2022}\natexlab{}.
\newblock \showarticletitle{An ensemble approach for spam detection in Arabic opinion texts}.
\newblock \bibinfo{journal}{\emph{Journal of King Saud University-Computer and Information Sciences}} \bibinfo{volume}{34}, \bibinfo{number}{1} (\bibinfo{year}{2022}), \bibinfo{pages}{1407--1416}.
\newblock


\bibitem[Sanh et~al\mbox{.}(2019)]%
        {sanh2019distilbert}
\bibfield{author}{\bibinfo{person}{Victor Sanh}, \bibinfo{person}{Lysandre Debut}, \bibinfo{person}{Julien Chaumond}, {and} \bibinfo{person}{Thomas Wolf}.} \bibinfo{year}{2019}\natexlab{}.
\newblock \showarticletitle{DistilBERT, a distilled version of BERT: smaller, faster, cheaper and lighter}.
\newblock \bibinfo{journal}{\emph{arXiv preprint arXiv:1910.01108}} (\bibinfo{year}{2019}).
\newblock


\bibitem[Saravanos et~al\mbox{.}(2021)]%
        {saravanos2021hidden}
\bibfield{author}{\bibinfo{person}{Antonios Saravanos}, \bibinfo{person}{Stavros Zervoudakis}, \bibinfo{person}{Dongnanzi Zheng}, \bibinfo{person}{Neil Stott}, \bibinfo{person}{Bohdan Hawryluk}, {and} \bibinfo{person}{Donatella Delfino}.} \bibinfo{year}{2021}\natexlab{}.
\newblock \showarticletitle{The hidden cost of using Amazon Mechanical Turk for research}. In \bibinfo{booktitle}{\emph{HCI International 2021-Late Breaking Papers: Design and User Experience: 23rd HCI International Conference, HCII 2021, Virtual Event, July 24--29, 2021, Proceedings 23}}. Springer, \bibinfo{pages}{147--164}.
\newblock


\bibitem[Sedoc and Ungar(2020)]%
        {sedoc2020item}
\bibfield{author}{\bibinfo{person}{Jo{\~a}o Sedoc} {and} \bibinfo{person}{Lyle Ungar}.} \bibinfo{year}{2020}\natexlab{}.
\newblock \showarticletitle{Item Response Theory for Efficient Human Evaluation of Chatbots}. In \bibinfo{booktitle}{\emph{Proceedings of the First Workshop on Evaluation and Comparison of NLP Systems}}. \bibinfo{pages}{21--33}.
\newblock


\bibitem[Shah and Bhavsar(2022)]%
        {shah2022time}
\bibfield{author}{\bibinfo{person}{Bhoomi Shah} {and} \bibinfo{person}{Hetal Bhavsar}.} \bibinfo{year}{2022}\natexlab{}.
\newblock \showarticletitle{Time complexity in deep learning models}.
\newblock \bibinfo{journal}{\emph{Procedia Computer Science}}  \bibinfo{volume}{215} (\bibinfo{year}{2022}), \bibinfo{pages}{202--210}.
\newblock


\bibitem[Tao et~al\mbox{.}(2018)]%
        {tao2018domain}
\bibfield{author}{\bibinfo{person}{Dapeng Tao}, \bibinfo{person}{Jun Cheng}, \bibinfo{person}{Zhengtao Yu}, \bibinfo{person}{Kun Yue}, {and} \bibinfo{person}{Lizhen Wang}.} \bibinfo{year}{2018}\natexlab{}.
\newblock \showarticletitle{Domain-weighted majority voting for crowdsourcing}.
\newblock \bibinfo{journal}{\emph{IEEE transactions on neural networks and learning systems}} \bibinfo{volume}{30}, \bibinfo{number}{1} (\bibinfo{year}{2018}), \bibinfo{pages}{163--174}.
\newblock


\bibitem[Tao et~al\mbox{.}(2020)]%
        {tao2020label}
\bibfield{author}{\bibinfo{person}{Fangna Tao}, \bibinfo{person}{Liangxiao Jiang}, {and} \bibinfo{person}{Chaoqun Li}.} \bibinfo{year}{2020}\natexlab{}.
\newblock \showarticletitle{Label similarity-based weighted soft majority voting and pairing for crowdsourcing.}
\newblock \bibinfo{journal}{\emph{Knowledge \& Information Systems}} \bibinfo{volume}{62}, \bibinfo{number}{7} (\bibinfo{year}{2020}).
\newblock


\bibitem[Tsironi et~al\mbox{.}(2017)]%
        {tsironi2017analysis}
\bibfield{author}{\bibinfo{person}{Eleni Tsironi}, \bibinfo{person}{Pablo Barros}, \bibinfo{person}{Cornelius Weber}, {and} \bibinfo{person}{Stefan Wermter}.} \bibinfo{year}{2017}\natexlab{}.
\newblock \showarticletitle{An analysis of convolutional long short-term memory recurrent neural networks for gesture recognition}.
\newblock \bibinfo{journal}{\emph{Neurocomputing}}  \bibinfo{volume}{268} (\bibinfo{year}{2017}), \bibinfo{pages}{76--86}.
\newblock


\bibitem[Vaswani et~al\mbox{.}(2017)]%
        {46201}
\bibfield{author}{\bibinfo{person}{Ashish Vaswani}, \bibinfo{person}{Noam Shazeer}, \bibinfo{person}{Niki Parmar}, \bibinfo{person}{Jakob Uszkoreit}, \bibinfo{person}{Llion Jones}, \bibinfo{person}{Aidan~N. Gomez}, \bibinfo{person}{Lukasz Kaiser}, {and} \bibinfo{person}{Illia Polosukhin}.} \bibinfo{year}{2017}\natexlab{}.
\newblock \showarticletitle{Attention is All You Need}.
\newblock
\urldef\tempurl%
\url{https://arxiv.org/pdf/1706.03762.pdf}
\showURL{%
\tempurl}


\bibitem[Verma et~al\mbox{.}(2024)]%
        {verma2024auditing}
\bibfield{author}{\bibinfo{person}{Preetika Verma}, \bibinfo{person}{Kokil Jaidka}, {and} \bibinfo{person}{Svetlana Churina}.} \bibinfo{year}{2024}\natexlab{}.
\newblock \showarticletitle{Auditing Counterfire: Evaluating Advanced Counterargument Generation with Evidence and Style}.
\newblock \bibinfo{journal}{\emph{arXiv preprint arXiv:2402.08498}} (\bibinfo{year}{2024}).
\newblock


\bibitem[Veselovsky et~al\mbox{.}(2023)]%
        {veselovsky2023artificial}
\bibfield{author}{\bibinfo{person}{Veniamin Veselovsky}, \bibinfo{person}{Manoel~Horta Ribeiro}, {and} \bibinfo{person}{Robert West}.} \bibinfo{year}{2023}\natexlab{}.
\newblock \showarticletitle{Artificial artificial artificial intelligence: Crowd workers widely use large language models for text production tasks}.
\newblock \bibinfo{journal}{\emph{arXiv preprint arXiv:2306.07899}} (\bibinfo{year}{2023}).
\newblock


\bibitem[Waseem(2016)]%
        {waseem-2016-racist}
\bibfield{author}{\bibinfo{person}{Zeerak Waseem}.} \bibinfo{year}{2016}\natexlab{}.
\newblock \showarticletitle{Are You a Racist or Am {I} Seeing Things? Annotator Influence on Hate Speech Detection on {T}witter}. In \bibinfo{booktitle}{\emph{Proceedings of the First Workshop on {NLP} and Computational Social Science}}. \bibinfo{publisher}{Association for Computational Linguistics}, \bibinfo{address}{Austin, Texas}, \bibinfo{pages}{138--142}.
\newblock
\urldef\tempurl%
\url{https://doi.org/10.18653/v1/W16-5618}
\showDOI{\tempurl}


\bibitem[Xin and Inkpen(2020)]%
        {xincl}
\bibfield{author}{\bibinfo{person}{Weizhao Xin} {and} \bibinfo{person}{Diana Inkpen}.} \bibinfo{year}{2020}\natexlab{}.
\newblock \showarticletitle{[CL-Aff Shared Task] Detecting Disclosure and Support via Deep Multi-Task Learning}. In \bibinfo{booktitle}{\emph{Proceedings of the 3rd Workshop on Affective Content Analysis (AffCon 2020) co-located with Thirty-Fourth {AAAI} Conference on Artificial Intelligence {(AAAI} 2020), New York, USA, February 7, 2020}} \emph{(\bibinfo{series}{{CEUR} Workshop Proceedings}, Vol.~\bibinfo{volume}{2614})}. \bibinfo{publisher}{CEUR-WS.org}, \bibinfo{pages}{118--129}.
\newblock
\urldef\tempurl%
\url{http://ceur-ws.org/Vol-2614/AffCon20_session3_detecting.pdf}
\showURL{%
\tempurl}


\bibitem[Xu et~al\mbox{.}(2015)]%
        {xu2015argumentation}
\bibfield{author}{\bibinfo{person}{Junyi Xu}, \bibinfo{person}{Li Yao}, {and} \bibinfo{person}{Le Li}.} \bibinfo{year}{2015}\natexlab{}.
\newblock \showarticletitle{Argumentation based joint learning: A novel ensemble learning approach}.
\newblock \bibinfo{journal}{\emph{PloS one}} \bibinfo{volume}{10}, \bibinfo{number}{5} (\bibinfo{year}{2015}), \bibinfo{pages}{e0127281}.
\newblock


\bibitem[Yang et~al\mbox{.}(2019a)]%
        {yang2019let}
\bibfield{author}{\bibinfo{person}{Diyi Yang}, \bibinfo{person}{Jiaao Chen}, \bibinfo{person}{Zichao Yang}, \bibinfo{person}{Dan Jurafsky}, {and} \bibinfo{person}{Eduard Hovy}.} \bibinfo{year}{2019}\natexlab{a}.
\newblock \showarticletitle{Let’s make your request more persuasive: Modeling persuasive strategies via semi-supervised neural nets on crowdfunding platforms}. In \bibinfo{booktitle}{\emph{Proceedings of the 2019 Conference of the North American Chapter of the Association for Computational Linguistics: Human Language Technologies, Volume 1 (Long and Short Papers)}}. \bibinfo{pages}{3620--3630}.
\newblock


\bibitem[Yang et~al\mbox{.}(2023)]%
        {yang2023survey}
\bibfield{author}{\bibinfo{person}{Yongquan Yang}, \bibinfo{person}{Haijun Lv}, {and} \bibinfo{person}{Ning Chen}.} \bibinfo{year}{2023}\natexlab{}.
\newblock \showarticletitle{A survey on ensemble learning under the era of deep learning}.
\newblock \bibinfo{journal}{\emph{Artificial Intelligence Review}} \bibinfo{volume}{56}, \bibinfo{number}{6} (\bibinfo{year}{2023}), \bibinfo{pages}{5545--5589}.
\newblock


\bibitem[Yang et~al\mbox{.}(2019b)]%
        {DBLP:journals/corr/abs-1906-08237}
\bibfield{author}{\bibinfo{person}{Zhilin Yang}, \bibinfo{person}{Zihang Dai}, \bibinfo{person}{Yiming Yang}, \bibinfo{person}{Jaime~G. Carbonell}, \bibinfo{person}{Ruslan Salakhutdinov}, {and} \bibinfo{person}{Quoc~V. Le}.} \bibinfo{year}{2019}\natexlab{b}.
\newblock \showarticletitle{XLNet: Generalized Autoregressive Pretraining for Language Understanding}.
\newblock \bibinfo{journal}{\emph{CoRR}}  \bibinfo{volume}{abs/1906.08237} (\bibinfo{year}{2019}).
\newblock
\showeprint[arxiv]{1906.08237}
\urldef\tempurl%
\url{http://arxiv.org/abs/1906.08237}
\showURL{%
\tempurl}


\bibitem[Zimmerman et~al\mbox{.}(2018)]%
        {zimmerman2018improving}
\bibfield{author}{\bibinfo{person}{Steven Zimmerman}, \bibinfo{person}{Udo Kruschwitz}, {and} \bibinfo{person}{Chris Fox}.} \bibinfo{year}{2018}\natexlab{}.
\newblock \showarticletitle{Improving hate speech detection with deep learning ensembles}. In \bibinfo{booktitle}{\emph{Proceedings of the Eleventh International Conference on Language Resources and Evaluation (LREC 2018)}}.
\newblock


\end{thebibliography}

\appendix
\section*{Supplementary Material}

\section{Datasets}
Table~\ref{tab:diplomacyannotation} and Table~\ref{tab:popularityannotation} detail the acquired and semantic attributes of the CLAff-Diplomacy and CLAff-OffMyChest datasets respectively. Table~\ref{tab:counterfire} details the grammatical quality annotation that was used in the cross-cohort comparison.
\begin{table}[!ht]
\caption{CLAff-Diplomacy final label and auxiliary label information (acquired label is Deception).}
\label{tab:diplomacyannotation}
\begin{center}
\small
 \resizebox{1.0\textwidth}{!}{
\begin{tabular}{p{3cm}p{4cm}p{7cm}}
\hline
\textbf{Label} & \textbf{Elaboration} & \textbf{Example \newline Messages} \\ \hline
Deception & Whether or not the receiver thinks the speaker is lying & ``I think I can top Germanys offer if youre interested." \newline
``I want it to be clear to you that youre the ally I want."\\
\hline
Game Move & Actual or suggested game move & ``I am committed to supporting Munich holding" \newline ``Make sure you don't move Munich so that it can take my support." \\ \hline
Reasoning & Justify a move, guess what moves might happen next, or discuss a move that already happened & ``If you took Marseilles, I would be stronger against England" \newline ``France is a really good player, and he is no doubt working hard to get England to turn on you."\\ \hline
Share Information & Share information related to other players & ``France is a really good player, and he is no doubt working hard to get England to turn on you." \newline ``And anything that's bad for Russia right now is good for Austria."\\ \hline
Rapport & Build rapport through compliments, concerns, reassurances or apologies & ``I won't hold it against you" \newline ``I'm going to keep helping you as much as I can." \\ \hline
\end{tabular}
}
\end{center}
\end{table}

\begin{table}[!ht]
\caption{CLAff-OffMyChest final label and auxiliary label information (acquired label is Popularity).}
\label{tab:popularityannotation}
\begin{center}
\small
 \resizebox{1.0\textwidth}{!}{
\begin{tabular}{p{3cm}p{4cm}p{7cm}}
\hline
\textbf{Label} & \textbf{Elaboration} & \textbf{Example \newline Messages} \\ \hline
Popularity & Whether a comment's upvotes are greater than or equal to the overall average number of upvotes & ``I don't know who is telling you this is bad parenting, but they are just flat out wrong." \newline
``I encourage you to report it, (...) I know how awful it can be to be raped (...)"\\
\hline
Emotional disclosure & Mentions the author’s feelings. & ``I love her and all, but I can leave her." \newline ``You're the reason that me has died?" \\ \hline
Information disclosure & Contains at least some personal
information about the author. & ``Hey, I don't do the exclusive partner thing." \newline ``(Extra info: I was also constantly in trouble for hurting her feelings, at times I would be called a bitch."\\ \hline
Emotional support & Offers sympathy, caring, or encouragement & ``Cry as much and as long as you need." \newline ``I'm 100\% with you that the OP should not pursue sex."\\ \hline
Information support & Offers specific information,
practical advice, or suggesting a course of action & ``	
6 - Hire a CPA with a good reputation!" \newline ``Ask for it to stay with you for every morning. :)  I know how you feel." \\ \hline
\end{tabular}
}
\end{center}
\end{table}

\begin{table}[!ht]
\caption{Counterfire annotation (acquired label is Persuasion).}
\label{tab:counterfireannotation}
\begin{center}
\small
 \resizebox{1.0\textwidth}{!}{
\begin{tabular}{p{3cm}p{4cm}p{7cm}}
\hline
\textbf{Label} & \textbf{Elaboration} & \textbf{Example \newline Messages} \\ \hline
Grammatical quality &  The quality of being grammatically correct and easy to understand & 1 (Poor): The statement contains many grammatical errors and is difficult to understand. \newline
2 (Fair): The statement contains some grammatical errors that may affect clarity. \newline
3 (Good): The statement is generally grammatically correct, but may contain occasional errors. \newline
4 (Excellent): The statement is well-written and largely free of grammatical errors. \newline
5 (Flawless): The statement is flawless in its grammar and syntax.\newline\\
\hline
\end{tabular}
}
\end{center}
\end{table}

\subsection{Direct prediction}
The results for training classifiers to directly predict the acquired variables (Deception and Popularity) for the CLAff-Diplomacy and CLAff-OffMyChest datasets are in Table~\ref{tab:individualscoresdataset1}(a) and (b) respectively. 

\subsection{Metafeature variants}
The pairwise correlation among the average metafeature values for the CLAff-Diplomacy dataset is provided in Figure~\ref{fig:corr}.

\begin{figure*}[!ht]
    \centering
    \includegraphics[width = 0.8\textwidth]{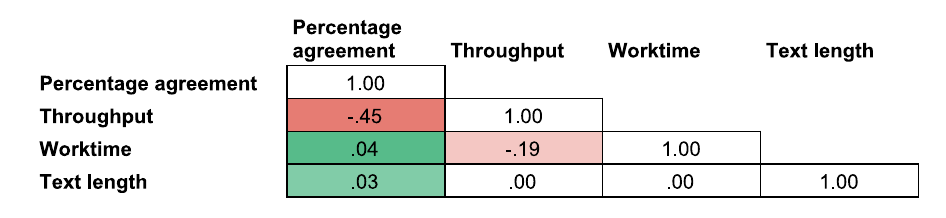}
    \caption{ Pairwise Pearson correlations between the average metafeature values. The cell shading reflects the direction and the magnitude of the correlation.}
    \label{fig:corr}
\end{figure*}

Metafeatures for the CLAff-OffMyChest dataset are in Figure~\ref{fig:popularitymetadataplots}.
Details of the metafeature variants are in Table~\ref{tab:encodinglayerweight}. 
\begin{table*}[!h]
\caption{The variations of weighting the Encoding Layer (labeled ``Probabilistic Encodings'' in Figure 1). ${a_j}_{1}^4$ are the encodings for the game move, reasoning, share information, and rapport models.} 
\label{tab:encodinglayerweight}
\resizebox{0.8\textwidth}{!}{%
\begin{tabular}{lr}
\hline
\textbf{Metadata} & \textbf{Weighting} \\ \hline
\textbf{Throughput} &  \\ 
TP1: average  $\overline{TP}$ & (TP1*a$_{1}$, TP1*a$_{2}$, TP1*a$_{3}$, TP1*a$_{4}$) \\ 
TP2: linear variance V(TP) & (TP2*a$_{1}$, TP2*a$_{2}$, TP2*a$_{3}$, TP2*a$_{4}$) \\
TP3: Negative quadratic variance & (TP3*a$_{1}$, TP3*a$_{2}$, TP3*a$_{3}$, TP3*a$_{4}$) \\
TP4: Quadratic variance & (TP4*a$_{1}$, TP4*a$_{2}$, TP4*a$_{3}$, TP4*a$_{4}$) \\
 \hline
\textbf{Worktime} & \\
WT1: average $\overline{WT}$ & (WT1*a$_{1}$, WT1*a$_{2}$, WT1*a$_{3}$, WT1*a$_{4}$) \\
WT2: linear variance V(WT) & (WT2*a$_{1}$, WT2*a$_{2}$, WT2*a$_{3}$, WT2*a$_{4}$) \\
\hline
\textbf{Percentage Agreement} & \\
PC1: average $\overline{PC}$ & (PC1*a$_{1}$, PC1*a$_{2}$, PC1*a$_{3}$, PC1*a$_{4}$) \\
PC2: linear variance V(PC) & (PC2*a$_{1}$, PC2*a$_{2}$, PC2*a$_{3}$, PC2*a$_{4}$) \\
PC3: original PC per model & (PC$_a$*a$_{1}$, PC$_b$*a$_{2}$, PC$_c$*a$_{3}$, PC$_d$*a$_{4}$) \\
\hline
\textbf{Text Length} & \\
TL1: normalised number of characters $\frac{chars}{max_{chars}}$ & (TL1*a$_{1}$, TL1*a$_{2}$, TL1*a$_{3}$, TL1*a$_{4}$)\\ 
TL2: normalised number of words $\frac{words}{max_{words}}$ & (TL2*a$_{1}$, TL2*a$_{2}$, TL2*a$_{3}$, TL2*a$_{4}$)\\
\hline
\textbf{Combination Options} & \\
SP1: 0.5*(TP1+TP2) & (SP1*a$_{1}$, SP1*a$_{2}$, SP1*a$_{3}$, SP1*a$_{4}$)\\
SP2: 0.5*(WT1+WT2) & (SP2*a$_{1}$, SP2*a$_{2}$, SP2*a$_{3}$, SP2*a$_{4}$)\\
SP3: 0.5*(PC1+PC2) & (SP3*a$_{1}$, SP3*a$_{2}$, SP3*a$_{3}$, SP3*a$_{4}$)\\
\hline
\end{tabular}
}
\end{table*}

\begin{figure*}[!ht]
    \centering
    \includegraphics[width=1.0\textwidth]{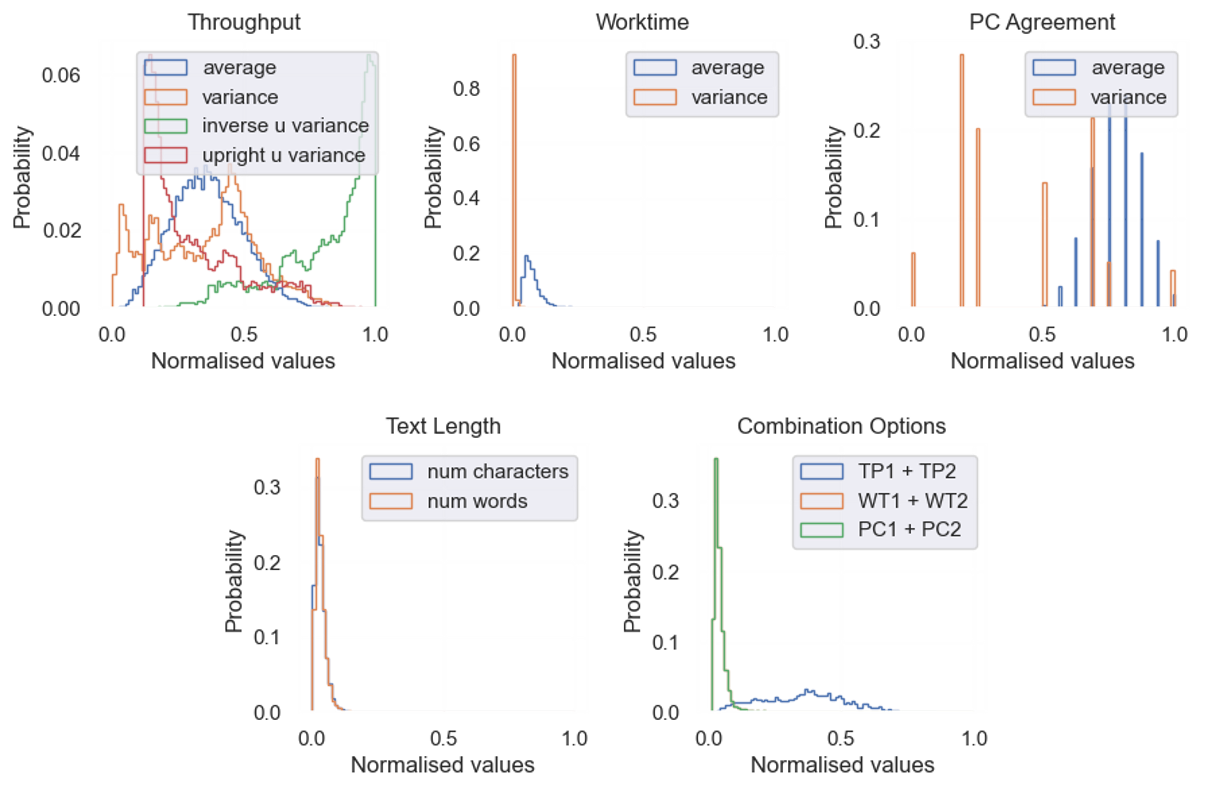} 
    \caption{Metadata Plots for CLAff-OffMyChest dataset}
    \label{fig:popularitymetadataplots}
\end{figure*}
 
\subsection{Full results}
The full results for applying all the models to the CLAff-Diplomacy dataset are provided in Table~\ref{tab:pipelinemodelsfull1} and ~\ref{tab:pipelinemodelsfull2}.
\begin{table}[!ht]
\caption{Full pipeline scores for Diplomacy dataset. Base model represents no weightage using AMT meta-features. The probabilistic encoding are joint using a Multi-Layer Perceptron. (Part 1)}
\label{tab:pipelinemodelsfull1}
\begin{center}
\begin{adjustbox}{totalheight=0.8\textheight}
 \resizebox{0.49\textwidth}{!}{%
\begin{tabular}{lrr}
\hline
\textbf{Model} & \multicolumn{2}{c}{\textbf{F1 Score}} \\ \hline
& \textbf{Rapport} & \textbf{Deception} \\ \hline
\multicolumn{3}{l}{\textbf{Baselines}} \\ 
MACE & 0.582 & - \\ 
Geva et. al & 0.474 & -\\ \hline
\multicolumn{3}{l}{\textbf{CNN}} \\ 
Base & 0.475 & \textbf{0.380} \\ 
TP1:  average $\overline{TP}$ & 0.114 & 0.377 \\ 
TP2:  linear variance $V(TP)$ & 0.456 & \textbf{0.445} \\
TP3:  quadratic variance $c<0$ &  0.500 & 0.173\\
TP4:  quadratic variance $c>0$ & 0.520 & 0.345 \\
WT1:  average $\overline{WT}$ & 0.475 & \textbf{0.487} \\
WT2:  linear variance $V(WT)$ & 0.210 & 0.487\\
PC1:  average $\overline{PC}$ & 0.431 & \textbf{0.553} \\
PC2:  linear variance $V(PC)$ & 0.576 & 0.341 \\
PC3:  original PC per model & 0.479 & 0.358 \\
TL1: normalised number of characters & 0.458 & 0.447 \\ 
TL2: normalised number of words & 0.510 & \textbf{0.518} \\
SP1: $0.5*(TP1+TP2)$ & 0.475 & \textbf{0.433} \\
SP2: $0.5*(WT1+WT2)$ & 0.466 & 0.056 \\
SP3: $0.5*(PC1+PC2)$ & 0.432 & 0.410 \\
\hline
\multicolumn{3}{l}{\textbf{LSTM}} \\ 
Base & 0.441 & \textbf{0.456} \\ 
TP1: average $\overline{TP}$ & 0.394 & \textbf{0.456} \\ 
TP2: linear variance $V(TP)$ & 0.351 & 0.351 \\
TP3: quadratic variance $c<0$ & 0.475 & 0.342 \\
TP4: quadratic variance $c>0$ & 0.486 & 0.311 \\
WT1: average $\overline{WT}$ & 0.475 & \textbf{0.494} \\ 
WT2: linear variance $V(WT)$ & 0.475 & 0.494 \\
PC1: average $\overline{PC}$ & 0.449 & \textbf{0.553} \\ 
PC2: linear variance $V(PC)$ & 0.463 & 0.374 \\
PC3: original PC per model & 0.415 & 0.469 \\ 
TL1: normalised number of characters & 0.478 & 0.395 \\
TL2: normalised number of words & 0.351 & \textbf{0.463} \\ 
SP1: $0.5*(TP1+TP2)$ & 0.542 & \textbf{0.379 }\\
SP2: $0.5*(WT1+WT2)$ & 0.313 & 0.069 \\
SP3: $0.5*(PC1+PC2)$ & 0.114 & 0.356 \\
\hline
\multicolumn{3}{l}{\textbf{LSTM-Attention}} \\ 
Base & 0.445 & \textbf{0.340} \\ 
TP1:  average $\overline{TP}$ & 0.489 & 0.095 \\
TP2:  linear variance $V(TP)$ & 0.535 & \textbf{0.444} \\ 
TP3: quadratic variance $c<0$ & 0.535 & 0.444 \\
TP4: quadratic variance $c>0$ & 0.496 & 0.374 \\
WT1: average $\overline{WT}$ & 0.437 & 0.329 \\
WT2: linear variance $V(WT)$ & 0.456 & \textbf{0.487} \\ 
PC1: average $\overline{PC}$ & 0.282 & 0.432 \\
PC2: linear variance $V(PC)$ & 0.483 & 0.417 \\
PC3: original PC per model & 0.477 & 0.414 \\ 
TL1: normalised number of characters & 0.447 & 0.114 \\
TL2: normalised number of words & 0.421 & \textbf{0.421} \\
SP1: $0.5*(TP1+TP2)$ & 0.537 & 0.421 \\
SP2: $0.5*(WT1+WT2)$ & 0.475 & \textbf{0.487} \\
SP3: $0.5*(PC1+PC2)$ & 0.475 & 0.047 \\ 
\hline
\multicolumn{3}{l}{\textbf{BiLSTM}} \\ 
Base & 0.434 & \textbf{0.326} \\ 
TP1:  average $\overline{TP}$ & 0.491 & \textbf{0.487} \\
TP2: linear variance $V(TP)$ & 0.491 & 0.487 \\ 
TP3: quadratic variance $c<0$ & 0.491 & 0.365 \\
TP4: quadratic variance $c>0$ & 0.489 & 0.270 \\
WT1: average $\overline{WT}$ & 0.464 & \textbf{0.329} \\
WT2: linear variance $V(WT)$ & 0.114 & 0.053 \\ 
PC1: average $\overline{PC}$ & 0.454 & \textbf{0.490} \\
PC2: linear variance $V(PC)$ & 0.454 & 0.490 \\
PC3: original PC per model & 0.469 & 0.456 \\ 
TL1: normalised number of characters & 0.466 & \textbf{0.476} \\
TL2: normalised number of words & 0.499 & 0.437 \\
SP1: $0.5*(TP1+TP2)$ & 0.508 & \textbf{0.441} \\
SP2: $0.5*(WT1+WT2)$ & 0.488 & 0.135 \\
SP3: $0.5*(PC1+PC2)$ & 0.458 & 0.089 \\ 
\hline
\end{tabular}
}
\end{adjustbox}
\end{center}
\end{table}
\begin{table}[!ht]
\caption{Full pipeline scores for Diplomacy dataset. Base model represents no weightage using AMT meta-features. The probabilistic encoding are joint using a Multi-Layer Perceptron. (Part 2)}
\label{tab:pipelinemodelsfull2}
\begin{center}
\begin{adjustbox}{totalheight=0.8\textheight}
 \resizebox{0.49\textwidth}{!}{%
\begin{tabular}{lrr}
\hline
\textbf{Model} & \multicolumn{2}{c}{\textbf{F1 Score}} \\ \hline
& \textbf{Rapport} & \textbf{Deception} \\ \hline
\multicolumn{3}{l}{\textbf{Baselines}} \\ 
MACE & 0.582 & - \\ 
Geva et. al & 0.474 & -\\ \hline
\multicolumn{3}{l}{\textbf{BERT}} \\ 
Base & 0.466 & \textbf{0.487} \\ 
TP1:  average $\overline{TP}$ & 0.310 & \textbf{0.466} \\ 
TP2: linear variance $V(TP)$ & 0.537 & 0.047 \\
TP3: quadratic variance $c<0$ &  0.466 & 0.047 \\
TP4: quadratic variance $c>0$ & 0.456 & 0.047 \\
WT1: average $\overline{WT}$ & 0.114 & 0.293 \\
WT2: linear variance $V(WT)$ & 0.480 & \textbf{0.487}\\
PC1: average $\overline{PC}$ & 0.466 & 0.362 \\
PC2: linear variance $V(PC)$ & 0.466 & 0.449 \\
PC3: original PC per model & 0.597 & \textbf{0.466} \\
TL1: number of characters & 0.508 & \textbf{0.047} \\ 
TL2: number of words & 0.467 & 0.047 \\
SP1: $0.5*(TP1+TP2)$ & 0.580 & 0.438 \\
SP2: $0.5*(WT1+WT2)$ & 0.466 & \textbf{0.487} \\
SP3: $0.5*(PC1+PC2)$ & 0.477 & 0.047 \\
\hline
\multicolumn{3}{l}{\textbf{Distilbert}} \\ 
Base & 0.466 & \textbf{0.487} \\ 
TP1: average $\overline{TP}$ & 0.528 & \textbf{0.420} \\ 
TP2: linear variance $V(TP)$ & 0.114 & 0.047 \\
TP3: quadratic variance $c<0$ & 0.114 & 0.189 \\
TP4: quadratic variance $c>0$ & 0466. & 0.299 \\
WT1: average $\overline{WT}$ & 0.457 &\textbf{0.332} \\
WT2: linear variance $V(WT)$ & 0.114 & 0.127\\
PC1: average $\overline{PC}$ & 0.417 & 0.114 \\
PC2: linear variance $V(PC)$ & 0.466 & 0.446 \\
PC3: original PC per model & 0.448 & \textbf{0.466} \\
TL1: number of characters & 0.114 & 0.384 \\ 
TL2: number of words & 0.531 & \textbf{0.421} \\
SP1: $0.5*(TP1+TP2)$ & 0.557 & 0.047\\
SP2: $0.5*(WT1+WT2)$ & 0.114 & \textbf{0.487} \\
SP3: $0.5*(PC1+PC2)$ & 0.482 & 0.487 \\
\hline
\multicolumn{3}{l}{\textbf{XLNet}} \\ 
Base & 0.114 & \textbf{0.487}\\ 
TP1: average $\overline{TP}$ & 0.114 & \textbf{0.487} \\ 
TP2: linear variance $V(TP)$ & 0.528 & 0.422 \\
TP3: quadratic variance $c<0$ & 0.417 & 0.282 \\
TP4: quadratic variance $c>0$ & 0.114 & 0.047 \\
WT1: average $\overline{WT}$ & 0.474 & \textbf{0.311} \\
WT2: linear variance $V(WT)$ & 0.114 & 0.104 \\
PC1: average $\overline{PC}$ & 0.114 & \textbf{0.418} \\
PC2: linear variance $V(PC)$ & 0.231 & 0.114 \\
PC3: original PC per model & 0.443 & 0.227 \\
TL1: number of characters & 0.499 & 0.047 \\ 
TL2: number of words & 0.466 & \textbf{0.487} \\
SP1: $0.5*(TP1+TP2)$ & 0.567 & \textbf{0.047} \\
SP2: $0.5*(WT1+WT2)$ & 0.523 & 0.047 \\
SP3: $0.5*(PC1+PC2)$ & 0.485 & 0.047 \\
\hline
\multicolumn{3}{l}{\textbf{RoBERTa}} \\ 
Base & 0.484 & \textbf{0.487} \\ 
TP1: average $\overline{TP}$ & 0.114 & 0.047\\ 
TP2: linear variance $V(TP)$ & 0.535 & 0.366 \\
TP3: quadratic variance $c<0$ & 0.447 & \textbf{0.487} \\
TP4: quadratic variance $c>0$ & 0.466 & 0.333 \\
WT1: average $\overline{WT}$ & 0.466 & 0.047 \\
WT2: linear variance $V(WT)$ & 0.466 & \textbf{0.487} \\
PC1: average $\overline{PC}$ & 0.399 & \textbf{0.487}\\
PC2: linear variance $V(PC)$ & 0.231 & 0.445 \\
PC3: original PC per model & 0.448 & 0.277 \\
TL1: number of characters & 0.514 & 0.411 \\ 
TL2: number of words & 0.560 & \textbf{0.487} \\
SP1: $0.5*(TP1+TP2)$ & 0.560 &\textbf{ 0.487} \\
SP2: $0.5*(WT1+WT2)$ & 0.516 & 0.487 \\
SP3: $0.5*(PC1+PC2)$ & 0.497 & 0.047 \\
\hline
\end{tabular}
}
\end{adjustbox}
\end{center}
\end{table}
The full results for applying all the models to the CLAff-OffMyChest dataset are provided in Table~\ref{tab:popularitytable}.
\begin{table}[!ht]
\caption{The best MSWEEM scores for the CLAff-OffMyChest dataset. Base model represents no weightage using AMT metadata features. The probabilistic encoding are joint using a Multi-Layer Perceptron.}
\label{tab:popularitytable}
\begin{center}
\begin{adjustbox}{totalheight=0.7\textheight-2\baselineskip}
 \resizebox{0.7\textwidth}{!}{
\begin{tabular}{lr}
\hline
\textbf{Model} & \textbf{F1 Score} \\ \hline
& \textbf{Popularity} \\ \hline
\multicolumn{2}{l}{\textbf{CNN}} \\ 
Base & \textbf{0.313} \\ 
TP1:  average $\overline{TP}$ & 0.314\\ 
TP2:  linear variance $V(TP)$ & 0.466 \\
TP3:  Negative quadratic variance & 0.451\\
TP4: Quadratic variance & \textbf{0.495} \\
WT1:  average $\overline{WT}$ & \textbf{0.206} \\
WT2:  linear variance $V(WT)$ & 0.184 \\
PC1:  average $\overline{PC}$ & \textbf{0.496}\\
PC2:  linear variance $V(PC)$ & 0.343 \\
PC3:  original PC per model & 0.345 \\
TL1: normalised number of characters & 0.156\\ 
TL2: normalised number of words & \textbf{0.449} \\
SP1: $0.5*(TP1+TP2)$ & \textbf{0.496} \\
SP2: $0.5*(WT1+WT2)$ & 0.449 \\
SP3: $0.5*(PC1+PC2)$ & 0.335 \\
\hline
\multicolumn{2}{l}{\textbf{LSTM}} \\ 
Base & \textbf{0.445} \\ 
TP1:  average $\overline{TP}$ & 0.156\\ 
TP2:  linear variance $V(TP)$ & 0.410 \\
TP3:  Negative quadratic variance & \textbf{0.459}\\
TP4: Quadratic variance & 0.156 \\
WT1:  average $\overline{WT}$ & \textbf{0.223} \\
WT2:  linear variance $V(WT)$ & 0.185 \\
PC1:  average $\overline{PC}$ & \textbf{0.480}\\
PC2:  linear variance $V(PC)$ & 0.459 \\
PC3:  original PC per model & 0.160 \\
TL1: normalised number of characters & \textbf{0.449}\\ 
TL2: normalised number of words & 0.156 \\
SP1: $0.5*(TP1+TP2)$ & \textbf{0.457} \\
SP2: $0.5*(WT1+WT2)$ & 0.296 \\
SP3: $0.5*(PC1+PC2)$ & 0.380 \\
\hline
\multicolumn{2}{l}{\textbf{LSTM-Attention}} \\ 
Base & \textbf{0.285} \\ 
TP1:  average $\overline{TP}$ & 0.417\\ 
TP2:  linear variance $V(TP)$ & 0.441 \\
TP3:  Negative quadratic variance &  0.471\\
TP4: Quadratic variance & \textbf{0.475} \\
WT1:  average $\overline{WT}$ & 0.156 \\
WT2:  linear variance $V(WT)$ & \textbf{0.449} \\
PC1:  average $\overline{PC}$ & 0.420\\
PC2:  linear variance $V(PC)$ & \textbf{0.425} \\
PC3:  original PC per model & 0.282 \\
TL1: normalised number of characters & \textbf{0.449}\\ 
TL2: normalised number of words & 0.398 \\
SP1: $0.5*(TP1+TP2)$ & \textbf{0.390} \\
SP2: $0.5*(WT1+WT2)$ & 0.304 \\
SP3: $0.5*(PC1+PC2)$ & 0.260 \\
\hline
\multicolumn{2}{l}{\textbf{BiLSTM}} \\ 
Base & \textbf{0.441} \\ 
TP1:  average $\overline{TP}$ & 0.500\\ 
TP2:  linear variance $V(TP)$ & 0.494 \\
TP3:  Negative quadratic variance & 0.445\\
TP4: Quadratic variance & \textbf{0.510} \\
WT1:  average $\overline{WT}$ & \textbf{0.223} \\
WT2:  linear variance $V(WT)$ & 0.189 \\
PC1:  average $\overline{PC}$ & 0.345\\
PC2:  linear variance $V(PC)$ & \textbf{0.439} \\
PC3:  original PC per model & 0.266 \\
TL1: normalised number of characters & \textbf{0.449}\\ 
TL2: normalised number of words & 0.449 \\
SP1: $0.5*(TP1+TP2)$ & 0.265 \\
SP2: $0.5*(WT1+WT2)$ & 0.392 \\
SP3: $0.5*(PC1+PC2)$ & \textbf{0.454} \\
\hline
\end{tabular}
}
\end{adjustbox}
\end{center}
\end{table}

\end{document}